\documentclass[10pt, logo, onecolumn]{shengshu}

\usepackage[utf8]{inputenc}
\usepackage[T1]{fontenc}
\usepackage{charter}

\newcommand{\bffont}{\fontsize{10}{10}\selectfont}
\DeclareTextFontCommand{\textbf}{\bffont\bfseries\selectfont}

\usepackage{wrapfig}
\usepackage{xspace}
\usepackage{textgreek}
\usepackage{bm}
\usepackage{cleveref}
\usepackage{multirow}
\usepackage{subcaption}
\usepackage{orcidlink}
\usepackage{footmisc}
\usepackage{algorithm}
\usepackage{algpseudocode}
\usepackage{listings}
\usepackage{csquotes}
\usepackage{marvosym}
\usepackage{nicefrac}
\usepackage{multicol}
\usepackage{tabto}
\usepackage{adjustbox}
\usepackage{dblfloatfix}
\usepackage{verbatim}
\usepackage{mathtools}
\usepackage{array}
\usepackage{bbm}
\usepackage{makecell}
\usepackage{siunitx}
\usepackage{pdflscape}
\usepackage{arydshln}
\usepackage{hhline}
\usepackage{diagbox}
\usepackage[square,sort,comma,numbers]{natbib}
\usepackage{fp}
\usepackage{tikz}
\usetikzlibrary{arrows.meta,positioning,fit,shapes.geometric,calc}

\newcolumntype{L}[1]{>{\raggedright\arraybackslash}p{#1}}

\definecolor{linkc}{rgb}{0, 0.44, 0.74}
\definecolor{eqc}{rgb}{1, 0, 0}
\definecolor{newcitecolor}{rgb}{0,0.6,0}
\definecolor{mygreen}{RGB}{34,139,34}
\definecolor{mylightblue}{RGB}{0,162,230}
\definecolor{deepyellow}{RGB}{255, 215, 0}
\definecolor{catgray}{gray}{0.92}
\definecolor{pearDark}{RGB}{171, 195, 87}
\definecolor{codebg}{RGB}{245, 245, 245}
\definecolor{keywordcolor}{RGB}{0, 0, 153}
\definecolor{commentcolor}{RGB}{34, 139, 34}
\definecolor{stringcolor}{RGB}{163, 21, 21}
\definecolor{numbercolor}{RGB}{128, 128, 128}

\newcommand{\ours}{Motus2\xspace}

\newcommand{\bx}{\boldsymbol{x}}

\def\onedot{\futurelet\@let@token\@onedot}
\def\@onedot{\ifx\@let@token.\else.\null\fi\xspace}

\makeatletter
\def\blfootnote#1{\xdef\@thefnmark{}\@footnotetext{\scriptsize #1}}
\makeatother

\hypersetup{
    colorlinks=true,
    urlcolor=shengshublue,
}

\let\cite\citep

\usepackage{amsmath,amsfonts,bm}

\def\eqref#1{equation~\ref{#1}}
\def\1{\bm{1}}

\DeclareMathAlphabet{\mathsfit}{\encodingdefault}{\sfdefault}{m}{sl}
\SetMathAlphabet{\mathsfit}{bold}{\encodingdefault}{\sfdefault}{bx}{n}

\title{
\fontsize{14}{22}\selectfont
Motus2: A Self-Evolving General World Model for Dexterous Manipulation
}

\author{
\begin{minipage}{\textwidth}
\centering
{\linespread{1.08}\selectfont
Hongzhe Bi$^{1,2\*\dagger}$,
Zihao Zhou$^{1\*}$,
Yihang Tang$^{1\*}$,
Jingrui Pang$^{1,2\*}$,
Shuhe Huang$^{1,2\*}$,
Haitian Liu$^{1,2}$,
Runqing Wang$^{1}$,
Shuai Huang$^{1}$,
Yichen Wang$^{1}$,
Yiming Cheng$^{2}$,
Ruowen Zhao$^{1,2}$,
Zhenghua Li$^{2}$,
Hengkai Tan$^{1,2}$,
Xiaolong Liu$^{1}$,
Jinhui Wan$^{1}$,
Jiabao Liu$^{1}$,
Min Zhao$^{1,2}$,
Fan Bao$^{1}$,
Jun Zhu$^{1,2\textrm{\Letter}}$
\par}
\vspace{0.5em}
$^{1}$GensPI \quad
$^{2}$Tsinghua University
\par\vspace{0.5em}
$^{\dagger}$Project lead. \quad
$^{\*}$Joint first authors. \quad
$^{\textrm{\Letter}}$Corresponding author.
\par\vspace{0.35em}
{\normalfont\small\texttt{bhz24@mails.tsinghua.edu.cn; dcszj@mail.tsinghua.edu.cn}}
\end{minipage}
}

\begin{abstract}
General embodied agents should perceive, predict, act, evaluate, and improve within a unified system. World models have shown great promise in building such agents, yet existing models typically append an action output head to a world simulator, without coupling them into a closed decision-and-learning loop for policy improvement. We present \textbf{Motus2}, a self-evolving general world model for dexterous manipulation. Motus2 advances world modeling through model scaling and data scaling. For \textbf{model scaling}, a single model with shared weights exposes three control interfaces: a policy (\emph{world--action model}), a simulator (\emph{action-conditioned world model}), and an evaluator (\emph{value model}). The policy proposes candidate action chunks, the simulator predicts their visual consequences, and the evaluator assesses the predicted outcomes. Their coupling forms a closed decision-and-learning loop for policy improvement. This formulation uses curated expert demonstrations for action learning, while failed and suboptimal interactions provide valuable evidence for dynamics modeling and value learning. For \textbf{data scaling}, Motus2 progresses from large-scale monocular egocentric data to synchronized stereo egocentric data, followed by robot-domain adaptation with robot trajectories and supplementary human--robot alignment data. Motus2 further studies global-autoregressive and hybrid-memory extensions of its sliding-window context, adds tactile feedback for contact-aware control, and is instantiated on a fully biomimetic platform with stereo vision, dual arms, dual dexterous hands, and tactile sensing. Together, egocentric data scaling and closed-loop general world model scaling
provide a general path toward self-evolving dexterous manipulation.

Project Page: \textbf{\href{https://motus-robotics.github.io/motus2}{\textcolor{shengshublue}{https://motus-robotics.github.io/motus2}}}.
\end{abstract}

\begin{document}
\maketitle

\section{Introduction}
\label{sec:introduction}

\nocite{bi2025motus,bi2026hrdt,diffusionnft2025,bao2023unidiffuser,
liu2024rdt,liu2026rdt2,motubrain2026,feng2025vidar,zhu2026gwm,bao2024vidu,
bao2023all,bao2022analytic,lu2022dpm,chen2023sfbc,
lu2023cep,chen2024srpo,chen2024eda,
ye2026worldactionmodelszeroshot,gao2026dreamdojo,egovla2025,
zheng2026egoscale,trex2026,dexmimicgen2025,
kim24openvla,black2024pi0,black2025pi05,
zhang2026dexora}

General-purpose dexterous manipulation requires more than a policy that maps
the latest observation to an action. A capable model should learn scalable
interaction priors, retain task-relevant history under self-occlusion, reason
about contact, and improve its policy beyond supervised imitation. Yet today's
robot foundation models are trained predominantly on curated action-supervised
datasets
~\cite{liu2024rdt,liu2026rdt2,kim24openvla,black2024pi0,black2025pi05,
feng2025vidar,chi2023diffusionpolicy}. Collecting embodiment-aligned robot
demonstrations at scale is expensive, while imitation alone provides no notion
of whether an action is good or bad and no mechanism for improving the policy
from its own outcomes. This motivates a self-evolving formulation in which the
model predicts and evaluates the consequences of candidate actions and uses
the resulting feedback for policy improvement.

Egocentric human data provides a scalable source of dexterous interaction
priors. Monocular egocentric data provide broad coverage of diverse tasks and
hand--object interactions
~\cite{grauman2022ego4d,egovla2025,zheng2026egoscale}, while synchronized
stereo egocentric data additionally provide implicit depth cues and support
more accurate 3D hand-pose estimation. However, transferring these priors to
dexterous robot execution remains challenging. Contact-critical events such as
fingertip slip, grasp formation, and release are often ambiguous from vision
alone, while first-person manipulation is partially observable because the
hands occlude objects and relevant consequences may appear only later. These
challenges motivate robot-domain adaptation, temporal context, and tactile
feedback
~\cite{trex2026,memorywam2026,hong2025relic}.

On the model side, recent control-oriented world models increasingly provide
two key capabilities. A \emph{world--action model} proposes executable actions
and can therefore serve as a policy, while an \emph{action-conditioned world
model} predicts their consequences and can therefore serve as a simulator.
Motus unifies these capabilities through UniDiffuser-style joint video--action
modeling within a single shared model
~\cite{bi2025motus,bao2023unidiffuser}. This reflects a key principle of
General World Models: action is not an auxiliary output attached to a
simulator, but the causal interface that grounds internal predictions in
physical interaction~\cite{zhu2026gwm}. However, predicted consequences alone
do not indicate whether an outcome advances the task. Self-evolution therefore
requires a third interface: an evaluator, realized by a value model, that
assesses predicted outcomes and provides signals for policy improvement. The
policy, simulator, and evaluator determine what action to attempt, what will
happen under that action, and whether the predicted outcome is desirable,
respectively. Therefore, the question is \emph{How to realize these causal interfaces within one
shared model without exposing future observations to action prediction, and to
convert outcome values into policy updates?}

To answer the above question, we present \ours, a \textbf{self-evolving General World
Model for dexterous manipulation} built on Motus. \ours scales dexterous manipulation along
two axes: \emph{data scaling} through a hierarchical egocentric human-data
pyramid and \emph{model scaling} through a  general world model with an explicit policy improvement closed-loop.
Specifically, for data scaling, pretraining progresses from large-scale monocular egocentric
data to synchronized stereo egocentric data, followed by robot-domain
mid-training on robot trajectories supplemented with human--robot alignment
data. For model scaling, one shared-parameter model exposes a policy
implemented by a world--action model, a simulator implemented by an
action-conditioned world model, and an evaluator implemented by a value model.
The policy proposes candidate action chunks, the simulator predicts their
visual consequences, and the evaluator estimates their task progress.
Best-of-$N$ planning uses these estimates for test-time selection, while
model-based reinforcement learning with
DiffusionNFT~\cite{diffusionnft2025} converts them into policy updates, closing
the decision-and-learning loop required for self-evolution.

This formulation separates imitation targets from other forms of interaction
evidence. Curated demonstrations supervise action learning, whereas failed and
suboptimal interactions provide evidence for dynamics
modeling and value learning. A lightweight tactile expert further supports
tactile-conditioned action refinement and tactile prediction for
contact-sensitive execution. To study long-horizon partial observability, we compare two extensions of the
bounded sliding-window context: global autoregression and hybrid memory~\cite{memorywam2026}.

In summary, our contributions are:
\begin{itemize}
    \item \textbf{A General World Model for dexterous manipulation}. \ours jointly
    models executable action chunks, action-conditioned future observations, and
    task-progress values within a shared-parameter video--action model. The model
    exposes three control interfaces: a policy implemented by a world--action
    model, a simulator implemented by an action-conditioned world model, and an
    evaluator implemented by a value model. A lightweight tactile expert further
    supports tactile-conditioned action refinement and tactile prediction for
    contact-sensitive execution
    (\S\ref{sec:method-overview}, \S\ref{sec:unified-world-model},
    \S\ref{sec:tactile-expert}).

    \item \textbf{An egocentric data-scaling and robot-domain transfer study}. Pretraining progresses from monocular egocentric data to
    stereo egocentric data, followed by robot-domain mid-training on robot
    trajectories and supplementary human--robot alignment data. Scaling
    experiments further establish a stereo human-data scaling trend
    (\S\ref{sec:unified-world-model}, \S\ref{sec:dataset},
    \S\ref{sec:human-data-scaling}).

    \item \textbf{A value-guided closed-loop self-evolution method}. The policy
    proposes candidate actions, the simulator predicts their consequences, and
    the evaluator estimates their task progress. Model-based reinforcement
    learning converts the resulting scores into policy updates, while the same
    interfaces support Best-of-$N$ test-time planning. Failed and suboptimal
    interactions provide evidence for dynamics modeling
    and value learning
    (\S\ref{sec:mbrl-planning}).
\end{itemize}

\section{Related Work}
\label{sec:related_work}

\paragraph{General World Models}
The term \emph{world model} covers systems with substantially different
representations and control capabilities, including latent dynamics for control,
generative prediction of observable futures, and interactive generation
conditioned on external inputs
~\citep{hafner2023dreamerv3,videoworldsimulators2024,bruce2024genie}. \emph{General World Model} (GWM) provide an overall framework for building a general foundation: it integrates an
understanding of the current world, imagination of possible and
action-conditioned futures, and action grounded by feedback from the external
world~\citep{zhu2026gwm}. We therefore distinguish the complete GWM from the
individual functional interfaces through which a model predicts, acts, or
evaluates outcomes.  
%a GWM further asks how these predictive capabilities connect to goal-directed action, outcome evaluation, and feedback.

Within robot control, we organize recent systems by three functional
interfaces. The \emph{policy} interface is realized by a \emph{world--action
model} (WAM), which generates actions from observations and task context.
Representative models such as Motus, DreamZero, Fast-WAM, MotuBrain, Being-H0.7, and Dyna-2 couple video modeling
with action generation for control
~\citep{bi2025motus,ye2026worldactionmodelszeroshot,fastwam2026,motubrain2026,
beingbeyond2026beingh07,dyna2026dyna2}. The \emph{simulator} interface is
realized by an \emph{action-conditioned world model}, which predicts future
observations under a proposed action; representative examples include
DreamDojo, Ctrl-World, and GigaWorld
~\citep{gao2026dreamdojo,ctrlworld2026,gigaworld2026}. The \emph{evaluator} interface is realized by a \emph{value model}~\citep{wang2026worldvalue}, reward model, or preference model that assesses
predicted outcomes.

In particular, Motus realizes the policy and simulator interfaces as conditional query modes
of one UniDiffuser-style video--action model, supporting video--action joint
prediction, VLA, inverse dynamics, video generation, and action-conditioned
future prediction with shared parameters
~\citep{bi2025motus,bao2023unidiffuser}. Motus2 develops this foundation for
dexterous manipulation by adding a value-based evaluator and model-based
reinforcement learning, thereby coupling policy, simulation, and evaluation
within one shared-parameter GWM. At a broader methodological level,
control-oriented world models build on advances in efficient diffusion sampling
and scalable generative architectures
~\citep{bao2022analytic,lu2022dpm,bao2023all,bao2024vidu}, as well as
diffusion-based behavior modeling and value-guided policy optimization
~\citep{chen2023sfbc,lu2023cep,chen2024srpo,chen2024eda,
diffusionnft2025}.

\paragraph{Dexterous Policies from Egocentric and Human Priors}

Dexterous manipulation requires high-DoF hands to coordinate discontinuous
contact dynamics, in which small pose or timing errors can alter the grasp
mode, induce slip, or prevent recovery. Recent systems address these challenges
through hand-specific action representations, retargeting, tactile-reactive
control, and scalable data generation, as in Dexora, DexMimicGen, T-Rex, and
METIS~\citep{zhang2026dexora,dexmimicgen2025,trex2026,metis2025}. Thus,
dexterity depends not only on policy capacity but also on embodiment-aware
supervision and contact feedback.

Egocentric human data provides scalable interaction priors across monocular
and stereo observation settings. Monocular egocentric data provides broad coverage of diverse
environments, objects, and human interaction behaviors, while stereo egocentric data additionally provides implicit depth cues from synchronized
binocular observations and more accurate 3D hand-pose estimates. H-RDT, EgoVLA,
and EgoScale demonstrate how human interaction data can strengthen robot policy
learning, while EgoSteer further leverages human--robot alignment for robot
execution~\citep{bi2026hrdt,egovla2025,zheng2026egoscale,egosteer2026}.
Being-H0.7 and EgoWAM add world-aware objectives over egocentric data
~\citep{beingbeyond2026beingh07,egowam2026}.

\paragraph{World-Model-Based Policy Improvement}

Model-based RL learns a dynamics model to predict future outcomes, thereby
providing supervision for policy improvement. Recent VLA post-training methods
extend beyond action imitation: WMPO optimizes policies with
world-model objectives, RISE uses a compositional world model for
self-improvement, Reinforcing Action Policies by Prophesying learns from
future-conditioned feedback, and NORA-1.5 constructs world-model and
action-based preference rewards~\citep{wmpo2025,rise2026,prophrl2025,
nora2026}. These methods show that predicted consequences can provide
supervision beyond expert demonstrations, especially when real robot
interaction is limited. An open question is how action decision-making, world simulation, and value
estimation can be coupled within a unified model to exploit the complementary
supervision provided by heterogeneous interaction data---including expert
demonstrations, successful trajectories, suboptimal executions, and
failures---for policy, dynamics, and value learning.

\section{Method}
\label{sec:method}

\subsection{Overview}
\label{sec:method-overview}

\begin{figure}[!t]
  \centering
  \captionsetup{font=small,skip=2pt}
  \vspace{-.15cm}
  \includegraphics[width=1.0\textwidth]{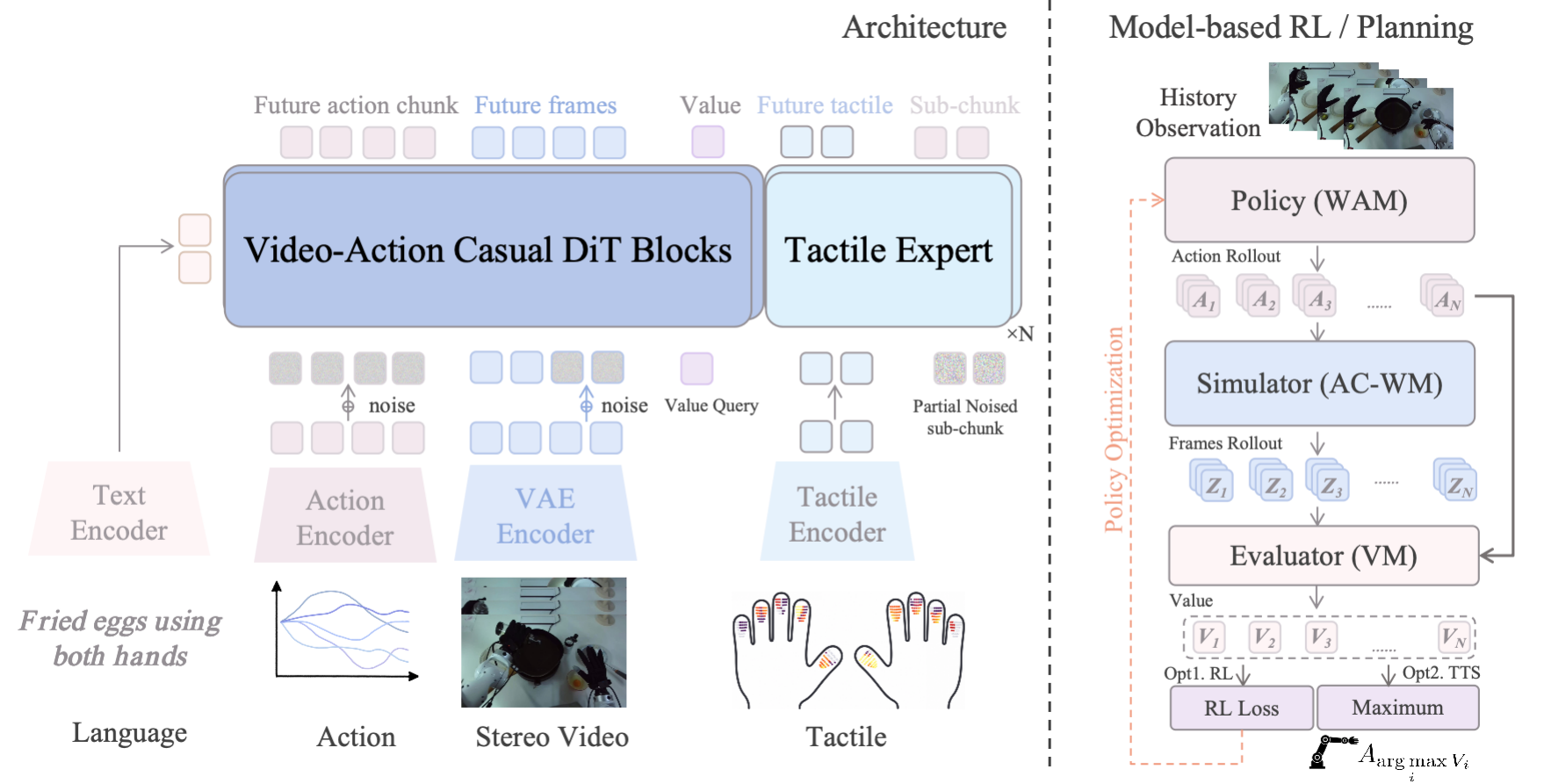}
  \vspace{-.1cm}
    \caption{\textbf{Motus2 overview.} A single general world model with one shared parameter set exposes three control interfaces. The policy generates action chunks from language, current robot observations, and the selected visual working-memory context; the simulator predicts future visual states under candidate actions; and the evaluator assesses the resulting branches for planning and policy optimization.}
  \label{fig:overall}
  \vspace{-.35cm}
\end{figure}

Fig.~\ref{fig:overall} summarizes \ours. We first learn a shared video--action
backbone through joint egocentric pre-training, then use action-first
factorization and trajectory-dependent supervision in robot-domain training to
expose policy, simulator, and evaluator interfaces from the same parameters.
Value-guided MBRL closes these interfaces into a learning loop, while working
memory and the tactile expert extend visual context and contact feedback,
respectively.

Formally, we consider the general task of language-conditioned dexterous
manipulation and formulate it as a partially observable Markov decision
process (POMDP):
\begin{equation}
    \mathcal{M}=\left(\mathcal{S},\mathcal{A},\mathcal{O},\mathcal{T},\Omega,r,\gamma\right),
    \label{eq:pomdp}
\end{equation}
where $s_t\in\mathcal{S}$ is the latent physical state,
$o_t=(I_t,q_t,\tau_t)\in\mathcal{O}$ contains vision, proprioception, and
optional tactile sensing, and $a_t\in\mathcal{A}$ is the robot action.
$\mathcal{T}$, $\Omega$, $r$, and $\gamma$ denote transition dynamics, the
observation model, reward, and discount. Because the hands frequently occlude
objects and contacts, the current observation is not a sufficient Markov state.
We therefore condition the deployed policy on the observation history
$h_t=(\ell,o_0,\ldots,o_t)$ under language instruction $\ell$. In practice,
the core model represents $h_t$ by a context $c_t$ containing language, current
proprioception, and an encoded visual history supplied by one of the
working-memory mechanisms in \S\ref{sec:working-memory}; the tactile expert
uses the concurrent tactile stream as described in \S\ref{sec:tactile-expert}.

Given this context, let $A_t=(a_t,\ldots,a_{t+H-1})$ be an executable action chunk,
$Z_t$ its latent future observations, and $Y_t$ a discretized task-progress
value. The support of $Y_t$ consists of the numerical centers of the progress
bins.
Starting at robot-domain mid-training, we factor their shared density in the
same action-first order used for control, planning, and policy optimization:
\begin{equation}
    p_\theta(A_t,Z_t,Y_t\mid c_t)
    =
    \underbrace{\pi_\theta(A_t\mid c_t)}_{\text{policy (WAM)}}
    \cdot
    \underbrace{p_\theta^{\mathrm{wm}}(Z_t\mid c_t,A_t)}_
    {\text{simulator (AC-WM)}}
    \cdot
    \underbrace{p_\theta^{\mathrm{vm}}(Y_t\mid c_t,A_t,Z_t)}_
    {\text{evaluator (VM)}}.
    \label{eq:action-first-factorization}
\end{equation}
These factors are three interfaces of the same shared-parameter model, not
separately trained architectures. The branch-ranking score is the evaluator's
conditional expectation
\begin{equation}
    V_\theta(c_t,A_t,Z_t)
    =
    \mathbb{E}_{Y_t\sim
    p_\theta^{\mathrm{vm}}(\cdot\mid c_t,A_t,Z_t)}
    \left[Y_t\right].
    \label{eq:value-expectation}
\end{equation}
This factorization defines the desired control chain; the remaining problem is
to realize it from a jointly pretrained video--action backbone without splitting
the three factors into separate networks.

\subsection{General World Modeling}
\label{sec:unified-world-model}
We address this challenge during robot-domain adaptation while preserving the
shared backbone. A stage-specific mask implements the policy--simulator--evaluator dependency in
Eq.~(\ref{eq:action-first-factorization}), while trajectory-dependent loss gates
route each trajectory only to the factors it can validly supervise.

\paragraph{Stage-specific information flow.}
Joint pre-training learns the video--action representation with mutual
within-chunk visibility and no value query. When robot-domain mid-training
begins, we introduce a read-only value query and organize each window as clean
teacher-forced observation history followed by action-first chunk blocks
\begin{equation}
    \bx=\left(Z^{\mathrm{ctx}};B_1;\ldots;B_M\right),
    \qquad
    B_j=\left(q_j;A_j;Z_j;U_j\right),
    \label{eq:uwm-sequence}
\end{equation}
where $q_j$ is the optional proprioception, $A_j$ is the action chunk, $Z_j$ is
its future video, and $U_j$ is a read-only value query. Action tokens cannot read
future-video or value tokens from their own chunk; future-video tokens may read
the current action; and the value query may read both but is hidden from all
other tokens. Across blocks, the mask is causal and windowed: block $j$ may use
preceding clean observations but never a later block. We call this design
\emph{chunk-autoregressive}: autoregression occurs across action chunks, while
the low-level actions within each chunk are generated jointly by flow matching.

\begin{figure}[t]
  \centering
  \begin{subfigure}[t]{0.47\linewidth}
    \centering
    \includegraphics[width=\linewidth]{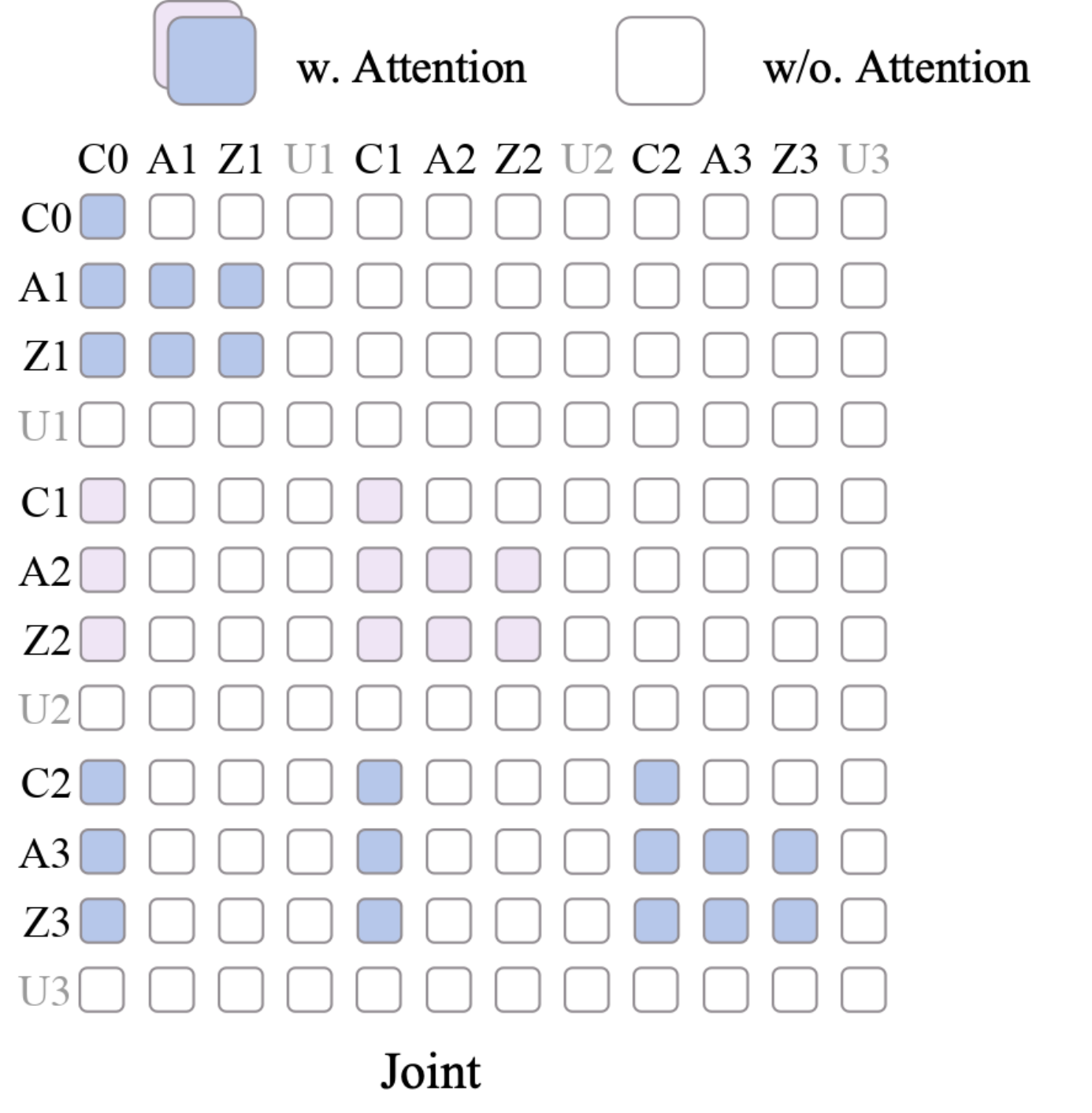}
    \caption{\footnotesize Joint mask for pre-training.}
  \end{subfigure}
  \hfill
  \begin{subfigure}[t]{0.47\linewidth}
    \centering
    \includegraphics[width=\linewidth]{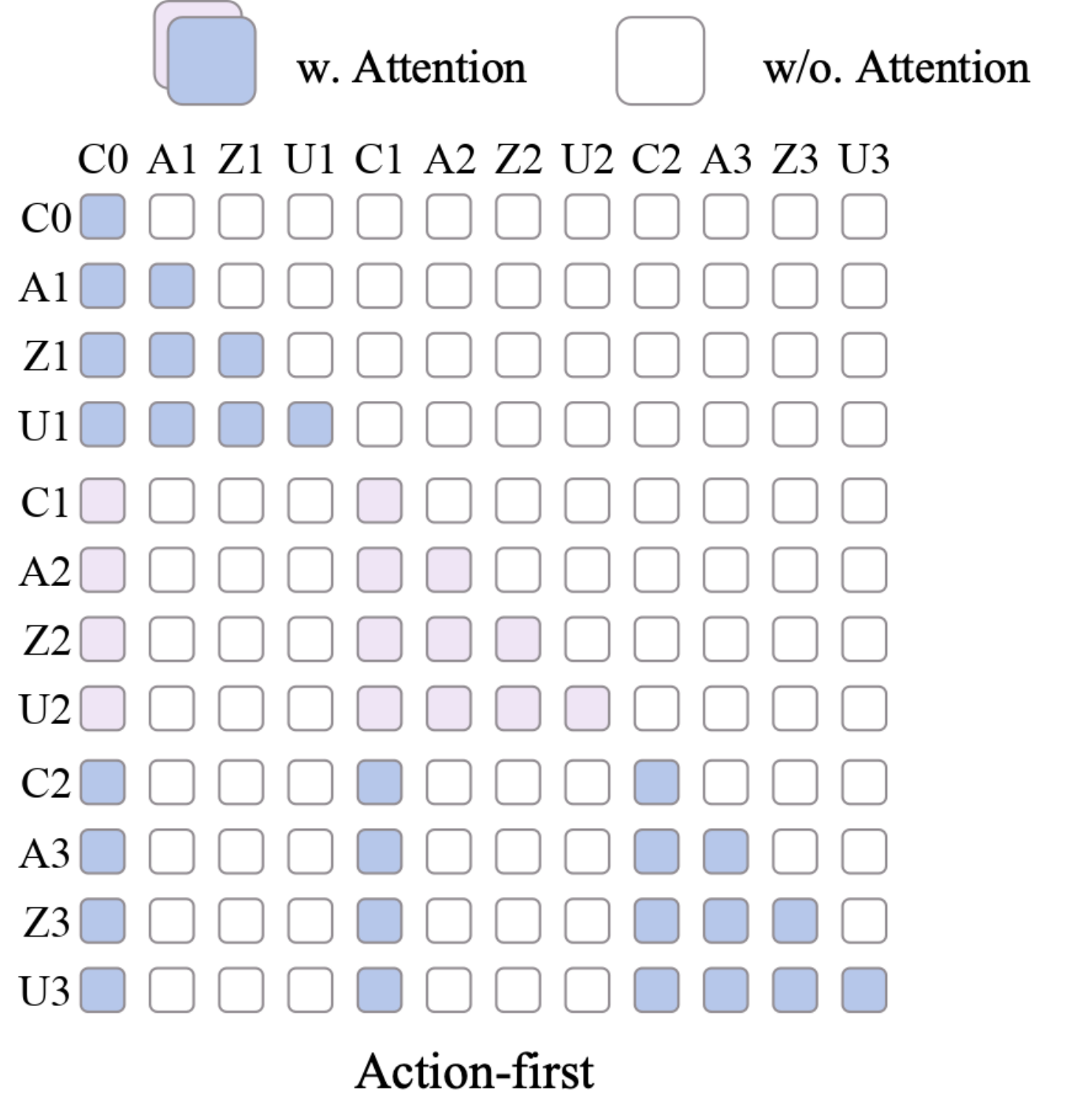}
    \caption{\footnotesize Action-first mask for mid or post-training.}
  \end{subfigure}
    \caption{\textbf{Stage-specific chunk masks.}
    Context ($C$), video ($Z$), and action ($A$) tokens share one backbone; a
    read-only value query ($U$) is introduced from mid-training onward and is
    inactive during joint pre-training. Joint pre-training permits bidirectional
    video--action interaction within a chunk. The action-first mask blocks current
    future-video information from action prediction, while future-video tokens may
    read the action and the value query may read both. Both masks remain causal
    across chunks.}
  \label{fig:post-training-masks}
\end{figure}

The resulting within-chunk dependency is
$A_j\rightarrow Z_j\rightarrow U_j$; tokenization, backbone, and cross-chunk
causality remain unchanged (Fig.~\ref{fig:post-training-masks}). Both stages use
the same backbone initialized from a foundation video diffusion
model~\cite{wan2025wan}. Stereo video latents share temporal and vertical
coordinates but occupy distinct horizontal RoPE ranges; proprioception, action,
and value queries enter the same transformer computation, while language enters
through cross-attention.

\paragraph{Trajectory-dependent supervision routing.}
The mask realizes the required dependency, but cannot determine whether a
recorded action should be imitated. Failed and suboptimal executions provide valid transition and
outcome evidence, but not desirable action targets. We therefore gate each
trajectory to the factors it can validly supervise. For a clean
target $x\in\{Z_j,A_j\}$ and Gaussian noise
$\epsilon\sim\mathcal{N}(0,I)$, we define
\begin{equation}
    x_\sigma=(1-\sigma)x+\sigma\epsilon,
    \qquad
    v^*=\epsilon-x,
    \qquad
    \sigma\in[0,1].
    \label{eq:uwm-flow}
\end{equation}
Our shared core predicts video and action velocity fields $v_\theta^z$ and
$v_\theta^a$, together with a categorical value distribution
$p_\theta^{\mathrm{vm}}$. We train them with the unified objective
\begin{equation}
    \mathcal{L}
    =
    w_z\left\|v_\theta^z-v_z^*\right\|_2^2
    +
    \lambda_a w_a\left\|v_\theta^a-v_a^*\right\|_2^2
    +
    \lambda_v w_v\,
    \mathrm{CE}\!\left(p_\theta^{\mathrm{vm}},Y_t\right),
    \label{eq:uwm-objective}
\end{equation}
where the loss gates $(w_z,w_a,w_v)$ and independent noise levels
$(\sigma^z,\sigma^a)$ select among three training modes under the action-first
mask:
\begin{equation}
\begin{array}{c|c|c|c}
\text{mode}
&
(\sigma^z,\sigma^a)
&
(w_z,w_a,w_v)
&
\text{supervised conditional}
\\ \hline
\text{policy}
&
(>0,>0)
&
(1,1,0)
&
p_\theta(A_t,Z_t\mid c_t)
\\
\text{simulation}
&
(>0,0)
&
(1,0,0)
&
p_\theta(Z_t\mid c_t,A_t)
\\
\text{evaluation}
&
(0,0)
&
(0,0,1)
&
p_\theta(Y_t\mid c_t,A_t,Z_t)
\end{array}
\label{eq:uwm-modes}
\end{equation}
From mid-training onward, policy mode jointly supervises action and future
video under the first two factors of
Eq.~(\ref{eq:action-first-factorization}), but deployment reads only its action
factor $\pi_\theta(A_t\mid c_t)$. Simulation mode keeps the recorded action
clean and supervises only action-conditioned future prediction, whereas
evaluation mode keeps action and video clean and supervises only the value
readout.

From mid-training onward, only curated successful trajectories activate action
supervision. We route failed and suboptimal trajectories, together with
task-irrelevant interactions, to the applicable simulation or evaluation mode,
where their recorded actions remain clean conditioning variables rather than
imitation targets. We thereby learn from their transitions and outcomes
without teaching the policy to reproduce undesirable behavior.

\paragraph{Data-scaling curriculum.}
The curriculum first learns joint human-interaction priors and introduces the
action-first mask and supervision routing only when robot-domain adaptation
begins. During
\emph{pre-training},
Stage~1 learns the video pathway from low- and then high-resolution monocular
egocentric clips using bidirectional conditional flow matching with one or two
clean latent frames as context. Stage~2 introduces synchronized stereo
egocentric data and jointly trains the left view, right view, and action tokens
with bidirectional within-chunk visibility. During \emph{mid-training}, we ground
these human priors in the robot domain using robot trajectories supplemented by
human--robot alignment data. At this point, we replace the joint interaction
with the action-first mask and mix the three modes in
Eq.~(\ref{eq:uwm-modes}). During \emph{post-training}, we retain this interface
for target-robot SFT and study three extensions built on it: value-guided MBRL,
long-history context, and tactile feedback, described in
\S\ref{sec:mbrl-planning}, \S\ref{sec:working-memory}, and
\S\ref{sec:tactile-expert}, respectively.

\paragraph{Action-first inference.}
Once trained, the factorization lets ordinary control query only the action factor
$\pi_\theta(A_t\mid c_t)$ and materialize future video and value only when
planning or policy optimization requires them. Ordinary control therefore stops
after the policy factor, whereas planning and policy improvement activate the
full policy--simulator--evaluator chain. This retains joint video--action
training without imposing a future rollout on every control
step~\cite{gigaworldpolicy2026,fastwam2026}.

\subsection{Value-Guided Closed-Loop Self-Evolution}
\label{sec:mbrl-planning}

\begin{figure}[H]
  \centering
  \begin{subfigure}[t]{0.49\linewidth}
    \centering
    \includegraphics[width=\linewidth]{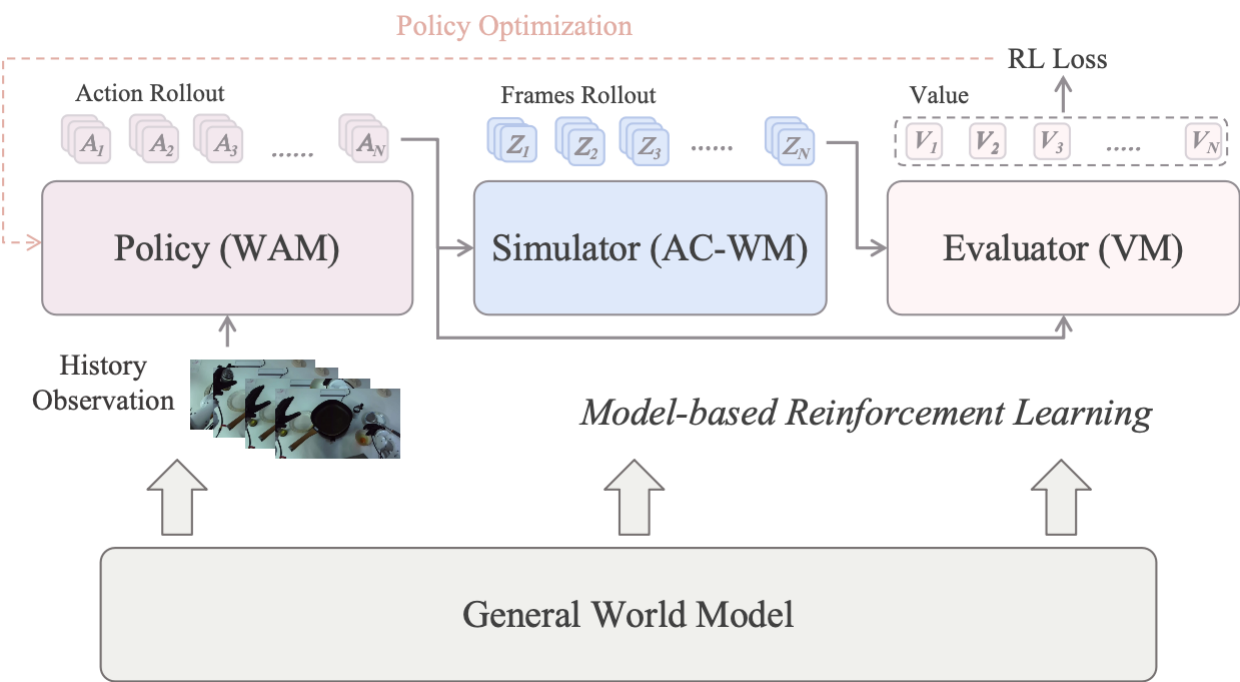}
    \caption{\footnotesize MBRL training loop.}
  \end{subfigure}
  \hfill
  \begin{subfigure}[t]{0.49\linewidth}
    \centering
    \includegraphics[width=\linewidth]{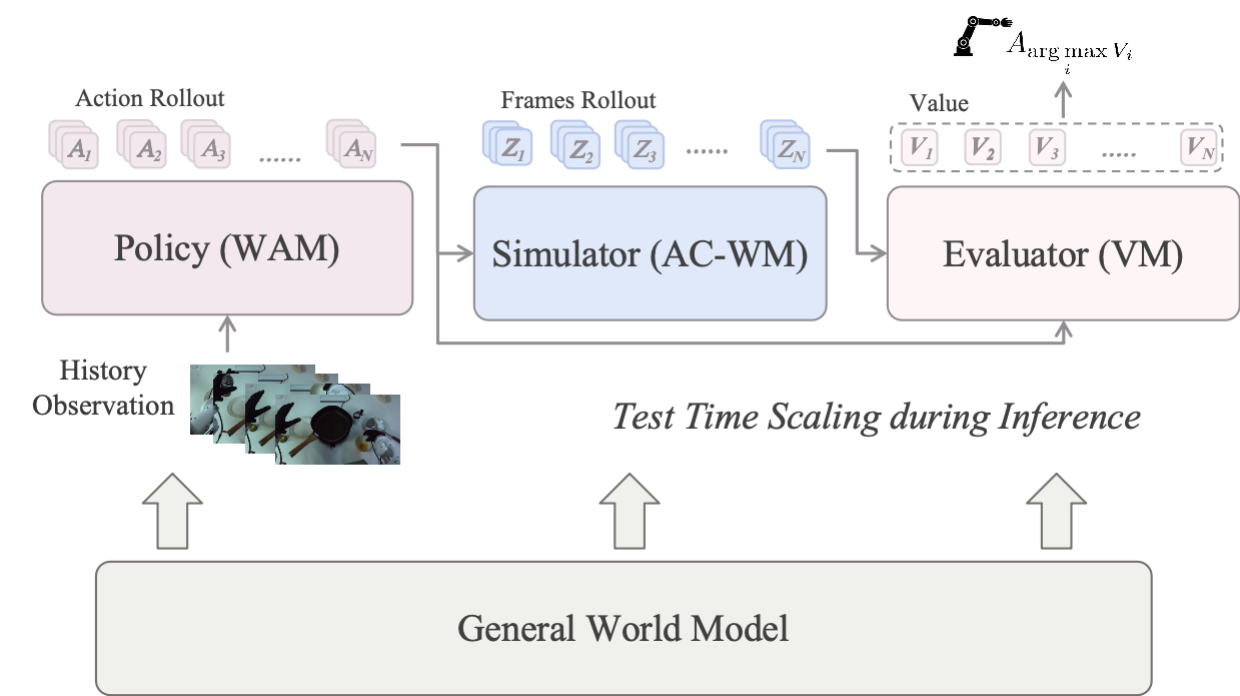}
    \caption{\footnotesize Test-time planning.}
  \end{subfigure}
\caption{\textbf{Self-evolution and planning modules.}
The Motus2 MBRL path samples action candidates with the world--action model
(policy), predicts their future visual consequences with the
action-conditioned world model (simulator), and evaluates the resulting
branches with the value model (evaluator). The resulting values guide
DiffusionNFT-style policy updates, while the same three interfaces support
test-time planning by selecting the highest-value branch.}
  \label{fig:mbrl-planning}
\end{figure}

The action-first chain makes candidate generation, simulation, and evaluation
executable in sequence, but self-evolution further requires the evaluator's
feedback to improve the policy. We train the evaluator with progress-based
supervision: successful trajectory segments provide positive progress targets,
whereas failed and task-irrelevant interactions provide negative supervision
for value learning. The resulting value estimates support both Best-of-$N$
selection and MBRL updates to the action pathway. Because the same
shared-parameter model realizes all three interfaces, \ours uses its own
predicted consequences and value estimates to improve its policy
(Fig.~\ref{fig:mbrl-planning}).

\subsubsection{Progress-Based Value Learning}
\label{sec:value-function-design}

To instantiate this progress-based supervision, we follow the
relative-progress formulation of VLAC~\cite{zhai2025vision} and assign each
segment from a successful trajectory the target
\begin{equation}
    r_t=\frac{\Delta t}{T-t},
    \label{eq:relative-progress-reward}
\end{equation}
where $\Delta t$ is the segment length, $T$ is the total trajectory length, and
$t$ is its starting timestep. The target lies in $[0,1]$ and measures the
progress of the current segment relative to the remaining task horizon. This
provides temporally localized supervision for evaluating action segments at
different stages of execution.

The positive targets above teach the evaluator how successful behavior advances,
but not how deviations from the task should be scored. We therefore complement
them with negative supervision from two sources. Failed teleoperation trajectories
capture realistic errors such as inaccurate manipulation, unintended contact,
and deviations from the instructed procedure. Task-irrelevant interactions
provide clear negative examples of behavior unrelated to the instructed task.

Following the relative-progress formulation, each negative trajectory segment is assigned
\begin{equation}
    r_t=-\frac{\Delta t}{T-t},
    \label{eq:random-negative-relative-reward}
\end{equation}
so negative segments represent behavior that does not contribute to task
completion. We discretize the resulting positive and negative rewards into the
categorical target $Y_t$ used by the evaluator in Eq.~(\ref{eq:uwm-objective}).

\subsubsection{Planning and Policy Optimization}
\label{sec:planning-policy-optimization}

With the evaluator defined above, the policy--simulator--evaluator chain
supports both action selection and policy improvement. At inference, value
estimates rank policy candidates for Best-of-$N$ planning; during MBRL, the
same estimates guide updates to the policy.

\paragraph{Model-Based Planning.}
At inference, we reuse the three interfaces as a receding-horizon planner:
propose $N$ chunks, simulate and score each branch, execute the highest-valued
chunk, append the real observation, and repeat. Replanning after each real
observation limits reliance on long imagined rollouts.
For branch $i$, the simulator predicts $Z_i$ and the evaluator returns
$V_i:=V_\theta(c_t,A_i,Z_i)$. We select
$i^\star=\arg\max_{i=1,\ldots,N}V_i$ and execute $A_{i^\star}$.

\paragraph{Policy Optimization.}
Planning improves selection but cannot change the proposal distribution. We
therefore feed evaluator scores back into the action pathway with
DiffusionNFT~\cite{diffusionnft2025}, which converts scalar branch scores into
a flow-matching policy update while keeping the simulator and evaluator fixed.

For every generated candidate pair $(A_i,Z_i)$ with value $V_i$, we sample a
noise level $\sigma_i$ and Gaussian noise $\epsilon_i$ and apply the forward
flow-matching process
\begin{equation}
    A_i^{\sigma_i}
    =(1-\sigma_i)A_i+\sigma_i\epsilon_i,
    \qquad
    v_i^{\mathrm{tar}}=\epsilon_i-A_i,
    \label{eq:nft-forward-noising}
\end{equation}
where $v_i^{\mathrm{tar}}$ is the flow-matching velocity target associated with the sampled clean action and noise.

The online policy and the EMA reference policy predict action velocity fields $v_{\theta,i}$ and $v_{\mathrm{ref},i}$ from the same noised action, noise level, language instruction, robot state, and clean observation history. The predicted future latent $Z_i$ affects the policy update through its value $V_i$, but is not exposed to the action-velocity predictor during NFT optimization.

We convert the scores of the policy-generated candidates into optimality weights:
\begin{equation}
    \hat r_i
    =
    0.5+0.5\cdot
    \operatorname{clip}
    \left(
    \frac{
    V_i-\operatorname{mean}(V)
    }{
    \max\!\left(\operatorname{std}(V),\varepsilon\right)
    },
    -1,
    1
    \right),
    \label{eq:reward-normalization}
\end{equation}
where $V=(V_1,\ldots,V_N)$, and $\operatorname{mean}(V)$ and
$\operatorname{std}(V)$ are the mean and population standard deviation of the
generated candidate values, respectively. The constant $\varepsilon>0$
prevents numerical instability when their variance is zero or near zero. The
resulting weight satisfies $\hat r_i\in[0,1]$: candidates valued above the mean
receive $\hat r_i>0.5$, whereas candidates valued below the mean receive
$\hat r_i<0.5$.

Each candidate group additionally contains one ground-truth action--future pair. It is assigned $\hat r_i=1$ as a positive behavioral anchor and is excluded from the mean and standard-deviation calculation over policy-generated candidates.

Following DiffusionNFT, we construct two implicit velocity fields around the reference prediction:
\begin{equation}
    v_i^{+}
    =(1-\beta)v_{\mathrm{ref},i}
    +\beta v_{\theta,i},
    \qquad
    v_i^{-}
    =(1+\beta)v_{\mathrm{ref},i}
    -\beta v_{\theta,i},
    \label{eq:nft-implicit-velocities}
\end{equation}
where the reference prediction is held fixed during each update. The coefficient $\beta>0$ controls the displacement of the implicit fields around the reference policy, with $1/\beta$ acting as the corresponding reinforcement-guidance strength.

The action policy is optimized using
\begin{equation}
    \mathcal{L}_{\mathrm{NFT}}
    =
    \mathbb{E}_i\!\left[
    \hat r_i
    \left\|v_i^{+}-v_i^{\mathrm{tar}}\right\|_2^2
    +
    \left(1-\hat r_i\right)
    \left\|v_i^{-}-v_i^{\mathrm{tar}}\right\|_2^2
    \right].
    \label{eq:diffusion-nft-loss}
\end{equation}
High-value candidates emphasize the positive implicit field, moving the online action distribution toward the corresponding actions. Low-value candidates emphasize the negative implicit field, moving the online distribution away from them. Candidates with $\hat r_i\approx0.5$ provide little directional preference and keep the online policy close to the reference.

Thus, planning changes selection from the current policy distribution, whereas
MBRL updates the policy distribution from which future candidates are drawn.

\paragraph{Asynchronous MBRL Training.}
The update above repeatedly executes candidate rollout, value scoring,
reference construction, and policy optimization. These stages have mismatched
throughput, so serial execution would leave expensive components idle. We
therefore implement a Ray-based macro-asynchronous pipeline, inspired by
AcceRL~\cite{accerl2026}, that decouples
offline prefix sampling, imagined rollout and value scoring, reference-target
construction, and policy optimization through bounded FIFO buffers. Rollout
actors generate candidate actions, predict their visual consequences, and
attach value scores. For each candidate, a reference actor samples the
flow-matching noise, constructs $A_i^{\sigma_i}$ and $v_i^{\mathrm{tar}}$, and
evaluates the same noised action with a versioned EMA policy to obtain the
fixed reference prediction $v_{\mathrm{ref},i}$. The trainer then evaluates
$v_{\theta,i}$ with the online policy and applies
Eq.~(\ref{eq:diffusion-nft-loss}).
Only action-related parameters are updated; the shared video backbone and
evaluator remain frozen. We tag samples with the EMA reference version and
discard stale samples or samples delayed beyond the configured bound.

\subsection{Working Memory}
\label{sec:working-memory}

The closed loop above can use only the evidence retained in $c_t$. A bounded
context may discard evidence observed early in a task before it is needed
again; hands may also hide the object in later views. We therefore make
accessible history an explicit design axis. A bounded sliding
window is our fixed-cost streaming default, and we implement two post-trained
long-history extensions---global autoregression and hybrid working
memory---that preserve earlier evidence in different forms.

\paragraph{Sliding window.}
We use fixed-length windows throughout joint pre-training and retain them under
the later action-first layout. From mid-training onward, each window contains
clean teacher-forced observations and the action-first prediction chunks
defined in \S\ref{sec:unified-world-model}. A prediction
chunk may attend to preceding clean observation chunks within the window, but
never to future chunks. An intermediate observation can therefore serve both
as the prediction target of its current chunk and, in clean form, as context
for later chunks. We vary the visible history width during training so that the
policy does not rely on a single fixed context length.

At deployment, we retain observed rather than imagined visual latents so the
context remains anchored to the executed trajectory. The KV cache keeps only
the most recent clean observations. After an action chunk is executed, the new
observation is appended and the oldest one is evicted when the cache is full. Temporal RoPE is
window-relative: after eviction, the retained video keys are rebased so that
their temporal coordinates remain within the positional range used in
training. Both cache capacity and per-step attention cost are therefore bounded
with respect to episode length, although observations that leave the window are
no longer accessible. The window settings are reported in
Appendix~\ref{app:model-training}.

\paragraph{Global autoregression.}
To remove the information loss caused by eviction, our global-autoregressive
variant retains all preceding clean visual latents. We use episode-level
temporal coordinates rather than rebasing positions within a window, allowing
each prediction chunk to attend to the full observation history. Its KV cache
and attention cost consequently grow with episode length.

\paragraph{Hybrid working memory.}
To preserve selected long-range evidence without retaining every old frame at
full resolution, we integrate a MemoryWAM-style hybrid
memory~\cite{memorywam2026}. It retains initial anchor frames and a recent
observation window at full resolution, while representing older intermediate
observations with persistent memory tokens.

\paragraph{Variable-length episode packing.}
The two long-history variants must learn across complete trajectories rather
than truncated windows, producing highly variable sequence lengths that make
fixed-count batching inefficient. We therefore train them on complete episodes
and pack those episodes by latent-frame load.

Specifically, we pack multiple episodes into one attention call using a
block-diagonal causal mask. No query can attend across episode boundaries;
rotary coordinates are generated independently for each episode, and padding
tokens contribute no loss. Packing therefore changes only the computational
layout and is equivalent to processing the episodes independently. The offline
packing plan balances total latent-frame load across micro-batches while
keeping every episode intact; an episode that exceeds the nominal frame budget
is processed as an untruncated singleton. The packing limits and optimization
settings are provided in Appendix~\ref{app:model-training}.

These mechanisms trade fixed-cost streaming, full-history retention, and
compressed persistence; we compare the two long-history extensions in
\S\ref{sec:memory-results}.

\subsection{Tactile Expert: Tactile Refinement and Prediction}
\label{sec:tactile-expert}

Working memory extends the available visual history, but vision may still leave
contact state ambiguous. To incorporate high-rate tactile evidence without
rerunning the full backbone, we introduce a
lightweight tactile expert that reuses the backbone's intermediate action chunk
and detached layer-wise key--value (KV) cache. Following T-Rex~\cite{trex2026},
the backbone denoises the full action chunk from $\sigma=1$ to a fixed
$\sigma_c$ once; immediately before each short sub-chunk is executed, the
expert uses the latest tactile window to complete its
$\sigma_c\rightarrow0$ update.

\paragraph{Tactile-conditioned action refinement.}
At decision time $t$, let $A_{t,k}^{\sigma_c}$ denote intermediate sub-chunk
$k$, $\boldsymbol{\tau}_{t,k}^{\mathrm{pre}}$ the tactile window immediately
preceding its execution, and
$\mathrm{KV}_{t}=\{\mathrm{KV}_{t}^{l}\}_{l=1}^{L}$ the cached visible-context
and refreshed action K/V. The tactile expert, parameterized by $\phi$, reads
$A_{t,k}^{\sigma_c}$, $\boldsymbol{\tau}_{t,k}^{\mathrm{pre}}$, and the
corresponding detached cache to predict the final action velocity
$v^{a}_{\phi,t,k}$. The refined sub-chunk is
\begin{equation}
    \widetilde A_{t,k}
    =A_{t,k}^{\sigma_c}-\sigma_c v^{a}_{\phi,t,k}.
    \label{eq:tactile-refinement}
\end{equation}
The same intermediate full action chunk and $\mathrm{KV}_{t}$ are reused across
the rolling updates; after one sub-chunk is executed, the newly acquired
tactile window conditions the next.

With $A_{t,k}$ denoting the clean action sub-chunk, the refinement loss is
\begin{equation}
    \mathcal{L}_{\mathrm{ref}}
    =\mathbb{E}_{t,k}
      \left[\left\|v^{a}_{\phi,t,k}
      -\frac{A_{t,k}^{\sigma_c}-A_{t,k}}{\sigma_c}
      \right\|_2^2\right].
    \label{eq:tactile-refinement-loss}
\end{equation}
The intermediate action and $\mathrm{KV}_{t}$ are detached when optimizing the
expert.

\paragraph{Tactile prediction.}
Action refinement uses the preceding tactile window to correct the next
sub-chunk. To also supervise how contact evolves under that action, we predict the real force window
$\mathbf{f}_{t,k}^{\mathrm{post}}$ following each action sub-chunk. Keeping the
preceding tactile window clean, we sample
$\boldsymbol{\epsilon}^{f}_{t,k}\sim\mathcal{N}(0,I)$ and
$\sigma^{f}_{t,k}\in[0,1]$ and apply the same flow-matching objective:
\begin{equation}
\begin{aligned}
    \mathbf{f}_{t,k}^{\sigma^f}
    &=(1-\sigma^{f}_{t,k})
      \mathbf{f}_{t,k}^{\mathrm{post}}
      +\sigma^{f}_{t,k}\boldsymbol{\epsilon}^{f}_{t,k},\\
    \mathcal{L}_{\mathrm{pred}}
    &=\mathbb{E}_{t,k,\sigma^{f},\boldsymbol{\epsilon}^{f}}
      \left[\left\|v^{f}_{\phi,t,k}
      -\left(\boldsymbol{\epsilon}^{f}_{t,k}
      -\mathbf{f}_{t,k}^{\mathrm{post}}\right)
      \right\|_2^2\right],
    \label{eq:tactile-prediction-loss}
\end{aligned}
\end{equation}
Here $v^{f}_{\phi,t,k}$ is decoded in normalized force-signal space. Future-force
tokens may attend to intermediate action tokens, whereas action-refinement
tokens cannot attend to the noised future-force tokens, preserving the
action-first dependency. For the Sharpa Wave, deformation images are clean
conditioning inputs only and are not predicted. The complete objective is
$\mathcal{L}_{\mathrm{tac}}=\mathcal{L}_{\mathrm{ref}}
+\lambda_{\mathrm{pred}}\mathcal{L}_{\mathrm{pred}}$. Force prediction is used
only during training; deployment outputs the refined actions. Architecture and
training settings are provided in
Appendix~\ref{app:model-training}.

\section{Large-Scale Egocentric Dataset}
\label{sec:dataset}
\label{sec:large_scale_ego_dataset}

We construct an approximately 130K-hour egocentric corpus from open-source
and procured data spanning monocular and stereo cameras, multiple resolutions,
diverse manipulation tasks, and varied environments.
Table~\ref{tab:ego_dataset_composition} reports approximate source-level raw
hours before quality filtering, temporal segmentation, and offline annotation.
The monocular portion provides broad visual and semantic coverage, while the
stereo portion additionally provides implicit depth cues and more accurate 3D
hand-pose estimates.

\begin{table}[!htbp]
\centering
\caption{Composition of the egocentric corpus. Report durations are approximate raw
recording hours before filtering, segmentation, and annotation.}
\label{tab:ego_dataset_composition}

\small
\setlength{\tabcolsep}{4pt}
\renewcommand{\arraystretch}{1.08}

\begin{tabular}{@{}lcccl@{}}
\toprule
Dataset & Camera Setup & Raw Hours & Resolution ($W\times H$) & Acquisition \\
\midrule
Egocentric-100K & Monocular & 100,500 & $480\times360$ & Open-source \\
Egocentric-10K  & Monocular & 10,000  & $480\times360$, $512\times512$ & Open-source \\
EgoVerse        & Monocular & 1,200   & $512\times512$ & Open-source \\
EgoDex          & Monocular & 800     & $480\times360$, $512\times512$ & Open-source \\
\midrule
Ropedia     & Stereo & 7,000 & $512\times512$ & Mixed \\
EgoScale    & Stereo & 6,000 & $640\times384$ & Procured \\
LightWheel  & Stereo & 1,200 & $640\times480$ & Procured \\
JD-Group    & Stereo & 2,000 & $512\times512$ & Procured \\
CyberOrigin & Stereo & 1,200 & $640\times416$ & Procured \\
\midrule
\textbf{Total} & Mixed & \textbf{130,000} & Mixed & Mixed \\
\bottomrule
\end{tabular}
\end{table}

Separate from the corpus above, we collect tens of hours of human--robot
alignment data using Wuji Human Gloves as supplementary mid-training data.
Mid-training uses more than 100 hours of data in total.
Collection details are provided in Appendix~\ref{app:human-robot-align}.

Pretraining follows the two-stage organization in
\S\ref{sec:unified-world-model}. Stage~1 uses low-resolution monocular data
for 500K steps and then high-resolution monocular data for 340K steps, while
Stage~2 introduces multi-resolution stereo egocentric data for 450K steps.
Compatible sources may contribute to multiple phases; for example, monocular
LightWheel recordings enter Stage~1, whereas its stereo recordings enter
Stage~2.

\begin{figure}[H]
  \centering
  \begin{subfigure}[t]{0.54\linewidth}
    \centering
    \includegraphics[width=\linewidth]{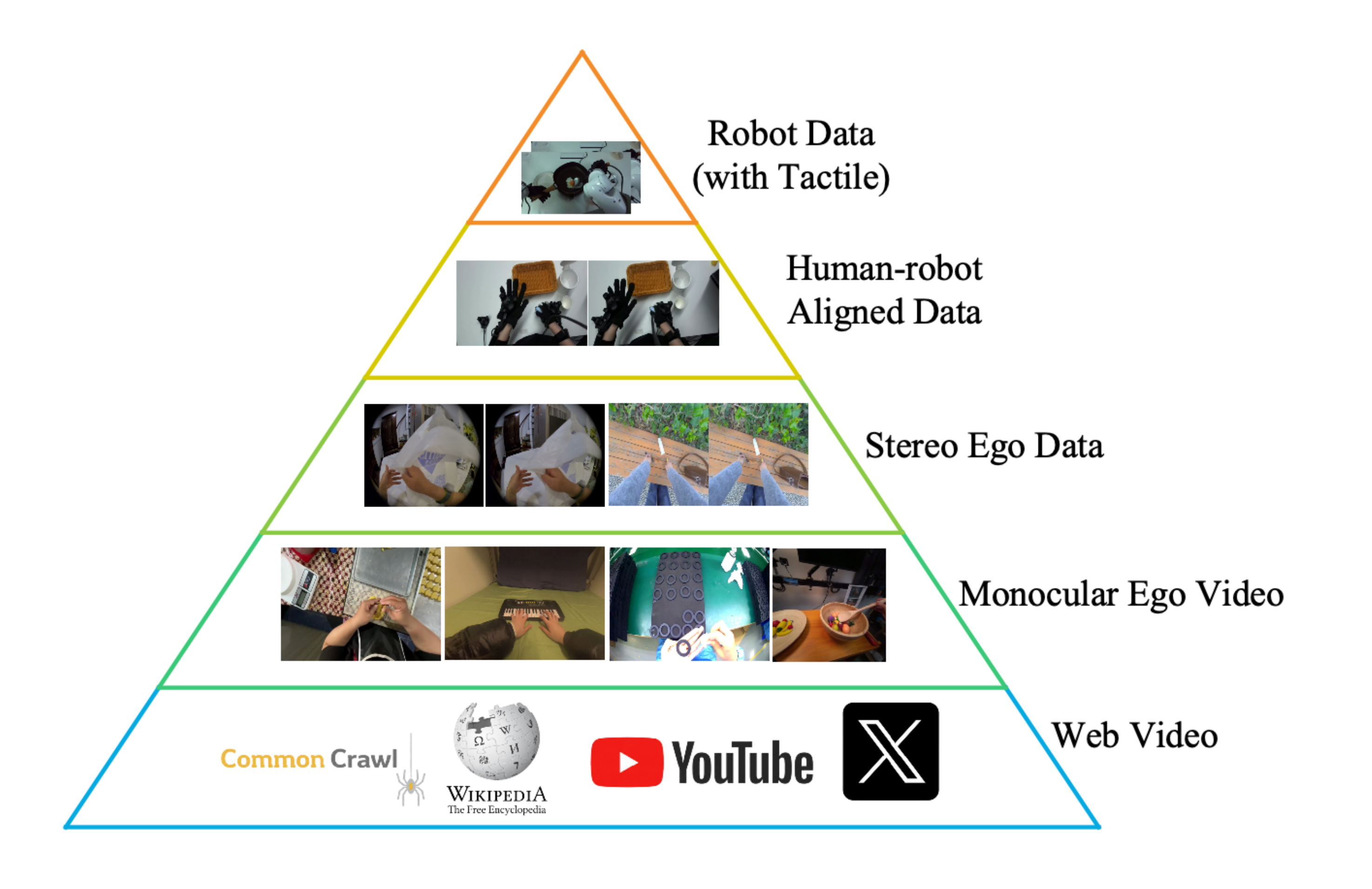}
  \end{subfigure}%
  \hspace{-0.03\linewidth}%
  \begin{subfigure}[t]{0.48\linewidth}
    \centering
    \IfFileExists{Figures/dataset_verb.pdf}{
      \includegraphics[width=\linewidth]{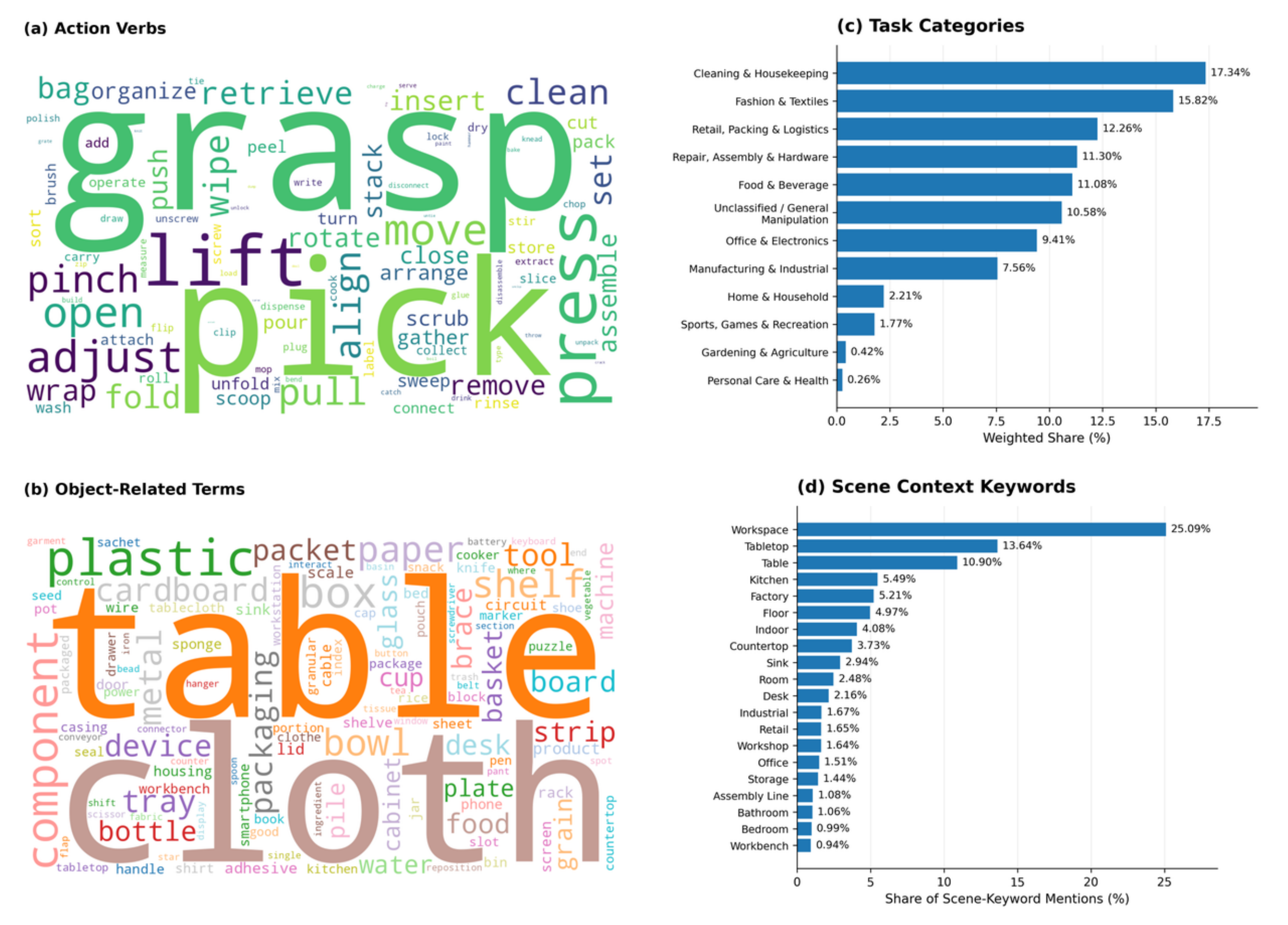}
    }{
      \fbox{\parbox[c][3.5cm][c]{0.94\linewidth}{\centering
      Semantic-diversity visualization placeholder.\\
      Add \texttt{Figures/dataset\_verb.pdf} to replace this box.}}
    }
  \end{subfigure}
  \caption{\textbf{Egocentric data pyramid and corpus semantics.}
  \textbf{Left:} the data hierarchy from web and monocular egocentric video
  to stereo egocentric data, human--robot alignment data, and robot
  interaction data. \textbf{Right:} semantic diversity of the annotated
  subset, excluding Egocentric-100K: \textbf{(a)} 99 fine-grained action
  verbs, \textbf{(b)} object-related terms, \textbf{(c)} 12 mutually
  exclusive high-level task categories, and \textbf{(d)} the top 20
  non-exclusive scene-context keywords.}
  \label{fig:egocentric-dataset-overview}
\end{figure}

The annotations span manipulation primitives, objects, tasks, and scenes.
Frequent actions include grasp, pick, lift, press, and adjust; task categories
cover cleaning, textiles, logistics, assembly, food preparation, office work,
manufacturing, household activities, recreation, gardening, and personal
care. All recordings are converted into synchronized episodes using a common
134-D hand-pose representation, quality-aware temporal segmentation, and
vision--language annotation. The complete processing protocol is provided in
Appendix~\ref{app:ego-data-processing}.

\section{Experiments}
\label{sec:experiments}

In this section, we aim to answer the following research questions through our experiments:

\begingroup
\setlength{\parskip}{0pt}
\noindent\textbf{RQ1:} \textit{How does egocentric human pretraining affect downstream robot control?}\par
\noindent\textbf{RQ2:} \textit{Do MBRL and test-time planning improve performance beyond supervised imitation?}\par
\noindent\textbf{RQ3:} \textit{How do different context mechanisms affect long-horizon control?}\par
\noindent\textbf{RQ4:} \textit{Does the tactile expert improve contact-rich manipulation?}\par
\endgroup

\subsection{Experiment Setup}
\label{sec:experiment-setup}

\paragraph{Task suites.}
The main suite contains five target-robot tasks spanning diverse control requirements: Place Ball and Put Phone test spatial grounding and placement;
Attach Eraser and Screw Bulb require precise object alignment and sustained
contact; and Multi-Finger tests coordinated multi-finger grasping and
manipulation of differently shaped objects.
The MBRL study uses Put Phone and Multi-Finger to isolate the effects of
policy optimization and test-time planning on the same physical tasks.
The memory study uses the separate Find Square and Press Button probes, whose
relevant evidence may be occluded or delayed beyond a short observation
window. This separation prevents the broad manipulation suite from obscuring
the specific effect of persistent context.
The tactile study uses Pull Out the Paper Cup and Tear Paper, which require
contact-sensitive grasp adjustment and sustained bimanual interaction.

\paragraph{Evaluation protocol.}
Within each evaluation setting, all methods use the same initial configurations
for each task. Real-robot evaluations use 20 configurations per task, whereas
the simulation branch of the long-horizon context study uses 25. Each method
executes one closed-loop rollout per configuration, using identical language
instructions, episode horizons, and task-specific success checkers.
A rollout is counted as successful only when all required task predicates are
satisfied within the horizon; partial completion is counted as failure. We
report success rate (SR), with the number of rollouts specified for each study.
The average is a macro-average over tasks, so each task receives equal weight.
Simulation and real-robot results are reported separately.

\paragraph{Controlled comparisons.}
We compare \ours against $\pi_{0.5}$~\cite{black2025pi05} under matched
target-task SFT data, observation interface, and evaluation protocol. WAN-SFT,
Pretrain-SFT, and \ours (Midtrain-SFT) use the same SFT procedure and differ
only in initialization: WAN, the two-stage egocentric-pretraining checkpoint,
and the robot-domain mid-trained checkpoint, respectively. Within each
controlled study, all other factors are held fixed.

\subsection{Main Results}
\label{sec:main-results}

\begin{table}[H]
  \centering
  \small
  \setlength{\tabcolsep}{3.5pt}
  \caption{\textbf{Main results.}
  Each entry reports success rate over 20 rollouts under matched target-robot
  finetuning. WAN-SFT, Pretrain-SFT, and Midtrain-SFT differ only in
  initialization.}
  \label{tab:main-results}
  \begin{tabular}{lcccccc}
    \toprule
    Method
      & Place Ball
      & Multi-Finger
      & Attach Eraser
      & Screw Bulb
      & Put Phone
      & Avg. SR $\uparrow$ \\
    \midrule
    $\pi_{0.5}$
      & 0\%
      & 0\%
      & 0\%
      & 0\%
      & 0\%
      & 0\% \\
    WAN-SFT
      & 0\%
      & 0\%
      & 0\%
      & 0\%
      & 0\%
      & 0\% \\
    Pretrain-SFT
      & 60\%
      & 35\%
      & 90\%
      & 55\%
      & 15\%
      & 51\% \\
    \ours (Midtrain-SFT)
      & \textbf{100\%}
      & \textbf{70\%}
      & \textbf{100\%}
      & \textbf{90\%}
      & \textbf{60\%}
      & \textbf{84\%} \\
    \bottomrule
  \end{tabular}
\end{table}

\paragraph{Results.}
Egocentric pretraining raises the macro-average success rate from 0\% for
WAN-SFT to 51\% for Pretrain-SFT. \ours (Midtrain-SFT) further increases the
average to 84\%, a 33-point gain over Pretrain-SFT. \ours reaches 100\% on Place
Ball and Attach Eraser, 90\% on Screw Bulb,
70\% on Multi-Finger, and 60\% on Put Phone. Under the matched target-task SFT
protocol, these comparisons support the contributions of egocentric
pretraining and subsequent robot-domain mid-training.

\subsection{Scaling Laws of Stereo Egocentric Human Data}
\label{sec:human-data-scaling}

Prior work has reported action-prediction scaling trends with monocular
egocentric human data~\cite{zheng2026egoscale,dyna2026dyna2}. We study the
corresponding trend for synchronized stereo observations and dexterous human
actions using nested 2k, 4k, 10k, and 20k-hour subsets. The subsets preserve
the source and resolution proportions, and all checkpoints are evaluated on
the same trajectory-disjoint stereo held-out set.

Held-out human-action prediction error serves as the measure of generalization. For each validation sample, the flow-matching policy generates multiple action chunks. Their mean prediction is compared with the normalized ground-truth wrist and hand actions using mean squared error. Let $\mathcal{L}_{\mathrm{val}}(D,e)$ denote the validation error after epoch $e$ for a model pretrained on $D$ hours of data. The optimal validation error at each data scale is
\begin{equation}
\mathcal{L}_{\mathrm{val}}^{*}(D)
\coloneqq
\min_e \mathcal{L}_{\mathrm{val}}(D,e).
\label{eq:human-data-scaling}
\end{equation}

Fitting the optimal error against data hours yields the log-linear relation
\begin{equation}
\mathcal{L}_{\mathrm{val}}^{*}(D)
\approx
0.101-0.005\ln D .
\label{eq:stereo-human-data-scaling-fit}
\end{equation}

\begin{figure}[t]
  \centering 
    \includegraphics[width=\linewidth]{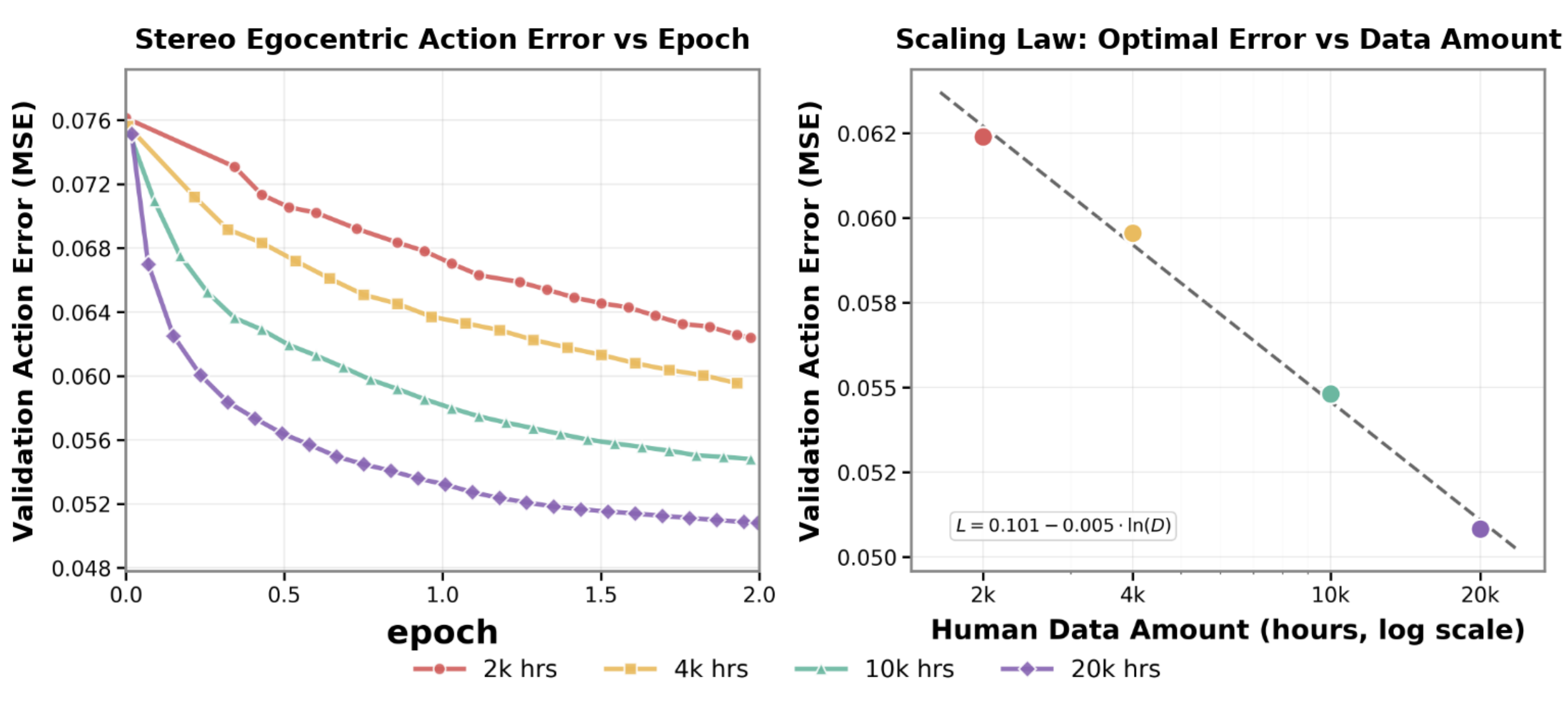}
  \caption{\textbf{Scaling laws of stereo egocentric human data.}
Left: held-out human-action prediction error across training epochs for models
trained on subsets constructed from 2k, 4k, 10k, and 20k raw recording hours
of stereo data. Right: optimal validation error versus raw stereo human-data
hours on a logarithmic scale, together with the fitted log-linear scaling law.
The reported hours are measured before quality filtering and temporal
segmentation; the effective amount of processed training data is therefore
smaller.}
  \label{fig:stereo-human-data-scaling}
\end{figure}

Figure~\ref{fig:stereo-human-data-scaling} presents the training dynamics and optimal validation errors. As the pretraining corpus grows from 2k to 20k hours, the validation curves shift consistently downward: smaller datasets plateau earlier, whereas larger datasets attain lower errors at comparable epochs. Across the four data scales, the optimal validation error decreases monotonically and is approximately linear in the logarithm of corpus size, indicating that the scaling trend previously observed for monocular egocentric data also extends to stereo observations within the measured range.

\subsection{Model-Based Policy Improvement and Test-Time Planning}
\label{sec:mbrl-results}

\paragraph{Factorial design.}
Starting from the same target-robot checkpoint, we independently enable
DiffusionNFT policy optimization and Best-of-$N$ planning. This yields a
$2\times2$ comparison of \ours, \ours + Planning, \ours + MBRL, and
\ours + MBRL + Planning.

\begin{table}[H]
  \centering
  \small
    \setlength{\tabcolsep}{6pt}
    \caption{\textbf{MBRL and planning results.}
    Each entry reports success rate over 20 rollouts. All variants start from the
    same target-robot checkpoint. MBRL updates the policy weights, whereas planning
    changes only inference.}
  \label{tab:mbrl-results}
  \begin{tabular}{lccc}
    \toprule
    Method
      & Put Phone
      & Multi-Finger
      & Avg. SR $\uparrow$ \\
    \midrule
    \ours
      & 60\%
      & 70\%
      & 65.0\% \\
    \ours + Planning
      & 65\%
      & 70\%
      & 67.5\% \\
    \ours + MBRL
      & 65\%
      & 80\%
      & 72.5\% \\
    \ours + MBRL + Planning
      & \textbf{70\%}
      & \textbf{80\%}
      & \textbf{75.0\%} \\
    \bottomrule
  \end{tabular}
\end{table}

\paragraph{Results.}
Planning raises \ours from 65.0\% to 67.5\% average success, showing a
2.5-point gain from online candidate selection without changing the policy
weights. \ours + MBRL reaches 72.5\% under direct inference, a 7.5-point gain
over \ours. Combining MBRL and planning yields 75.0\%. Planning therefore
remains beneficial after MBRL (2.5 points), while MBRL remains beneficial with
planning enabled (7.5 points), showing observed gains from policy optimization
and test-time selection in the evaluated tasks.

\paragraph{Qualitative value trajectories.}
Appendix~\ref{app:value-model-visualizations} visualizes predicted task
progress along one successful and three failed Screw Bulb trajectories. The
successful trajectory rises from $0.34$ to approximately $0.81$, whereas the
failed trajectories decrease after task progress is lost. These examples
illustrate the signal used for candidate ranking.

\subsection{Long-Horizon Context Mechanisms}
\label{sec:memory-results}

\paragraph{Protocol.}
On the Find Square and Press Button probes, we compare global autoregression
and hybrid working memory as described in \S\ref{sec:working-memory}. The two
variants use matched training and paired
evaluation configurations in simulation and on the real robot.

\begin{table}[H]
  \centering
  \small
  \setlength{\tabcolsep}{6pt}
  \caption{\textbf{Evaluation of long-horizon context mechanisms.}
  Simulation success rates are computed over 25 rollouts and real-robot rates
  over 20. Tasks are the long-horizon memory probes
  Find Square (\texttt{find\_the\_square}) and
  Press Button (\texttt{press\_button}).}
  \label{tab:memory-results}
  \begin{tabular}{llccc}
    \toprule
    Setting
      & Context mechanism
      & Find Square
      & Press Button
      & Avg. SR $\uparrow$ \\
    \midrule
    \multirow{2}{*}{Simulation}
      & Hybrid memory
      & 64\%
      & 40\%
      & 52\% \\
      & Global autoregression
      & 84\%
      & 72\%
      & 78\% \\
    \midrule
    \multirow{2}{*}{Real robot}
      & Hybrid memory
      & 30\%
      & 20\%
      & 25.0\% \\
      & Global autoregression
      & 65\%
      & 50\%
      & 57.5\% \\
    \bottomrule
  \end{tabular}
\end{table}

\paragraph{Results.}
Global autoregression outperforms hybrid memory on both probes in simulation,
reaching 84\% on Find Square and 72\% on Press Button, compared with 64\% and
40\% for hybrid memory. Its simulation macro-average is therefore 78\%, versus
52\% for hybrid memory. The same ordering holds on the real robot: global
autoregression achieves 65\% and 50\% on the two tasks, whereas hybrid memory
reaches 30\% and 20\%, giving macro-averages of 57.5\% and 25\%, respectively.
Global autoregression performs better than hybrid memory in both simulation
and real-robot evaluations.

\subsection{Effect of Tactile Feedback}
\label{sec:tactile-results}

\paragraph{Protocol.}
Both variants start from the same mid-trained checkpoint and use the same
target-task post-training data. The w/o Tactile variant removes the tactile
expert, whereas the w/ Tactile variant uses the complete refinement and
prediction objective in \S\ref{sec:tactile-expert}. We evaluate both variants on
Pull Out the Paper Cup with the Sharpa Wave and Tear Paper with Wuji Hand~2.

\begin{table}[H]
  \centering
  \small
  \setlength{\tabcolsep}{6pt}
  \caption{\textbf{Ablation of tactile feedback from the same mid-trained checkpoint.}
  Each entry reports the success rate over 20 rollouts.}
  \label{tab:tactile-results}
  \begin{tabular}{lccc}
    \toprule
    Variant
      & Pull Out Paper Cup
      & Tear Paper
      & Avg. SR $\uparrow$ \\
    \midrule
    w/o Tactile
      & 65\%
      & 55\%
      & 60.0\% \\
    w/ Tactile
      & 75\%
      & 70\%
      & 72.5\% \\
    \bottomrule
  \end{tabular}
\end{table}

\paragraph{Results.}
The tactile expert improves success on Pull Out the Paper Cup from 65\% to
75\% and on Tear Paper from 55\% to 70\%. The macro-average increases from
60.0\% to 72.5\%, a 12.5-point gain. These results indicate that tactile
feedback is beneficial for fine-grained, contact-sensitive manipulation in
the evaluated tasks.

\section{Conclusion}
\label{sec:conclusion}

We presented \ours, a self-evolving General World Model for dexterous manipulation. \ours treats action generation, action-conditioned future simulation, and outcome evaluation as different conditional functions of a shared physical-world model rather than as separate systems. With one shared set of weights, the model operates as a world--action model that proposes executable actions, an action-conditioned world model that predicts their visual consequences, and a value model that evaluates the predicted outcomes. Coupling these functions forms a closed decision-and-learning loop, enabling model-based policy improvement through candidate generation, consequence prediction, value-based selection, and policy updates.

On the data side, \ours treats egocentric human interaction as a scalable source of physical experience and progresses from large-scale monocular observation to stereo egocentric data and robot-domain grounding. The framework uses different interaction trajectories according to the supervision they provide: curated successful demonstrations supervise action learning, whereas failed and suboptimal interactions provide valuable evidence for learning dynamics and outcome evaluation. Tactile feedback further strengthens physical grounding during contact-sensitive manipulation.
Together, data scaling through egocentric human interaction and model scaling
through closed-loop world modeling provide a concrete path toward
self-evolving General World Models for dexterous manipulation.

\section{Limitations and Future Work}
\label{sec:limitations}

\paragraph{Limitations.}
The primary limitation of \ours is the scalability of wearable tactile supervision across embodiments. Whether worn by a human or a robot, a tactile glove deforms as the hand changes configuration. Even in the absence of external contact, material strain and internal fabric contact produce noisy tactile signals that are difficult to distinguish from meaningful physical contact. More fundamentally, current dexterous hands are not yet geometrically isomorphic to human hands: their little fingers are often elongated, and some are substantially thicker or larger than human hands. Consequently, a single glove pattern cannot fit both human and robotic hands, so tactile data collected with a human glove cannot be directly transferred to a dexterous hand wearing an embodiment-specific glove. This morphology-induced gap limits the scale and reliability of cross-embodiment tactile learning.

\paragraph{Future Work.}
From the perspective of \emph{data scaling}, advances in wearable tactile materials and fabrication could enable scalable, whole-hand tactile sensing for fully multimodal egocentric human data collection. Continued progress in anthropomorphic dexterous hand design could further narrow the morphological gap between robot and human hands, bringing robot interaction data closer to human data and enabling near-isomorphic cross-embodiment learning. From the perspective of \emph{model scaling}, stronger video foundation models and unified multimodal models can provide richer generative priors, longer-horizon predictive dynamics, and more faithful multimodal simulation. Building General World Models on these foundations could unify visual, language, action, tactile, and value signals within a common generative and decision-making process, predict the consequences of candidate actions over extended horizons, and continually improve from large-scale interaction.

\section*{Acknowledgments}

\paragraph{Hardware Acknowledgments.}
The robotic hands used in this work were commercially procured from \textit{WUJI} and \textit{SHARPA}, while the robotic arms and robotic platforms were commercially procured from \textit{Tianji}. We are especially grateful to \textit{Yunzhe Pan}, \textit{Jun Chen}, \textit{Zhen Wang} and \textit{Bo Liu} at \textit{WUJI} for their substantial technical support throughout the development of this project.

\paragraph{Data Acknowledgments.}
We acknowledge \textit{Ropedia}, \textit{EgoScale}, \textit{LightWheel}, \textit{JD-Group}, and \textit{CyberOrigin} as the providers of the commercially acquired datasets used in this work.

\section*{Core Contributors}

\noindent\textbf{Hongzhe Bi}: Base Model, Pre-Training, MBRL, Memory, Tactile Sensing, Sharpa \& WuJi-2 Hardware, Real Robot Post-Training, Ego Data, Writing.\par
\noindent\textbf{Zihao Zhou}: MBRL, Memory, Simulation, Real Robot Post-Training, Sharpa \& WuJi-2 Hardware, Writing.\par
\noindent\textbf{Yihang Tang}: Sharpa \& WuJi-2 Hardware, Tactile Sensing, Real Robot Post-Training.\par
\noindent\textbf{Jingrui Pang}: WuJi-1 Hardware, Human Data Collection, Real Robot Post-Training.\par
\noindent\textbf{Shuhe Huang}: MBRL, Simulation, Writing.\par

\clearpage

{
  \small
  \bibliographystyle{unsrt}
  \bibliography{main}
}

\clearpage
\appendix
\section{Robot System and Teleoperation Stack}
\label{app:robot-system}

\subsection{Robot Platforms}

\begin{figure}[H]
  \centering
  \captionsetup[subfigure]{justification=centering,singlelinecheck=false}
  \begin{subfigure}[t]{0.48\linewidth}
    \centering
    \parbox[c][0.20\textheight][c]{\linewidth}{\centering
      \includegraphics[height=0.20\textheight]
      {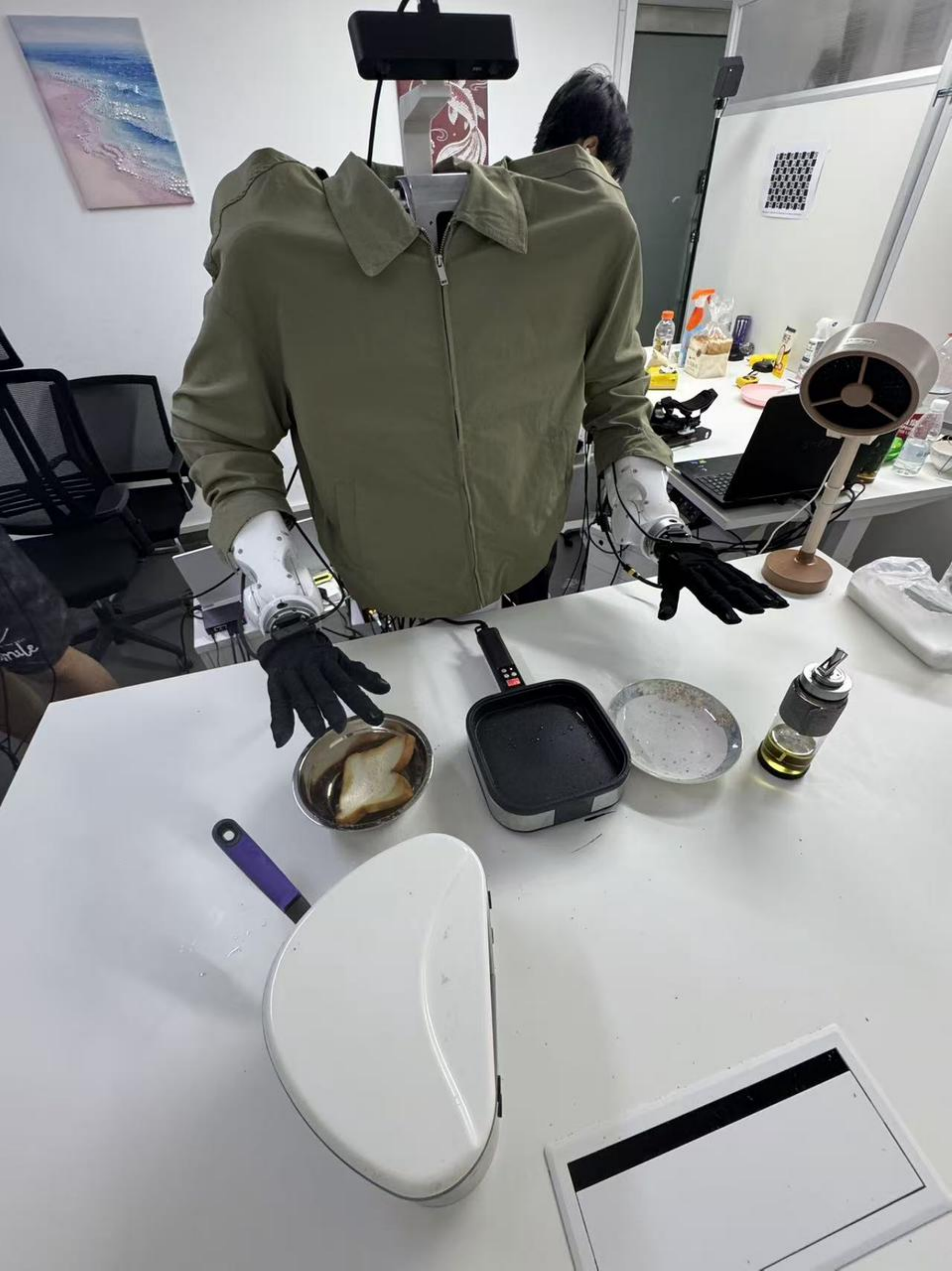}}
    \caption{\footnotesize Wuji hands with Tianji Marvin dual arms.}
  \end{subfigure}
  \hfill
  \begin{subfigure}[t]{0.48\linewidth}
    \centering
    \parbox[c][0.20\textheight][c]{\linewidth}{\centering
      \includegraphics[height=0.20\textheight]
      {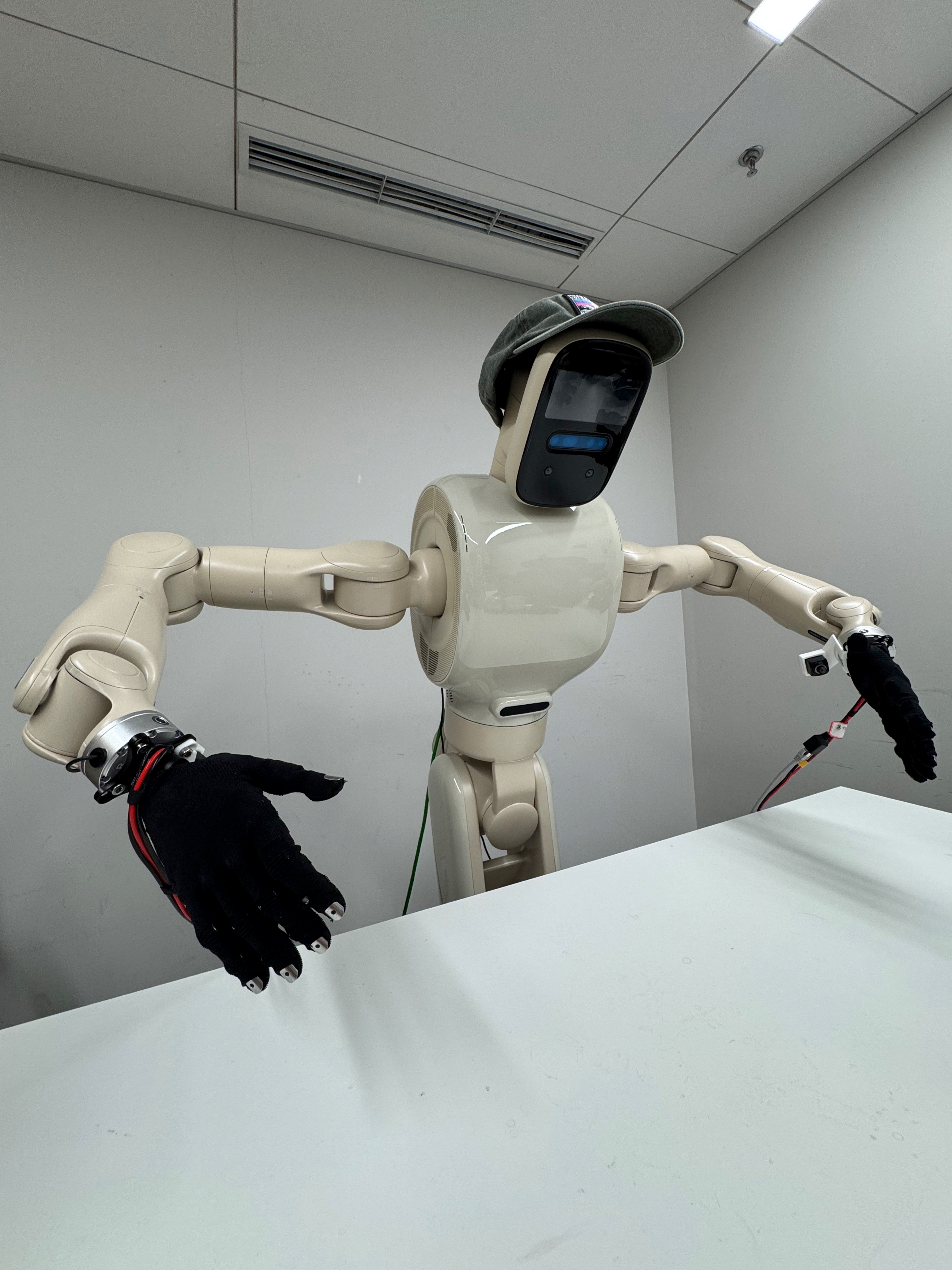}}
    \caption{\footnotesize Wuji Hand~2 with Tianji Gento Luna.}
  \end{subfigure}
  \par\smallskip
  \begin{subfigure}[t]{0.48\linewidth}
    \centering
    \parbox[c][0.20\textheight][c]{\linewidth}{\centering
      \includegraphics[height=0.20\textheight]
      {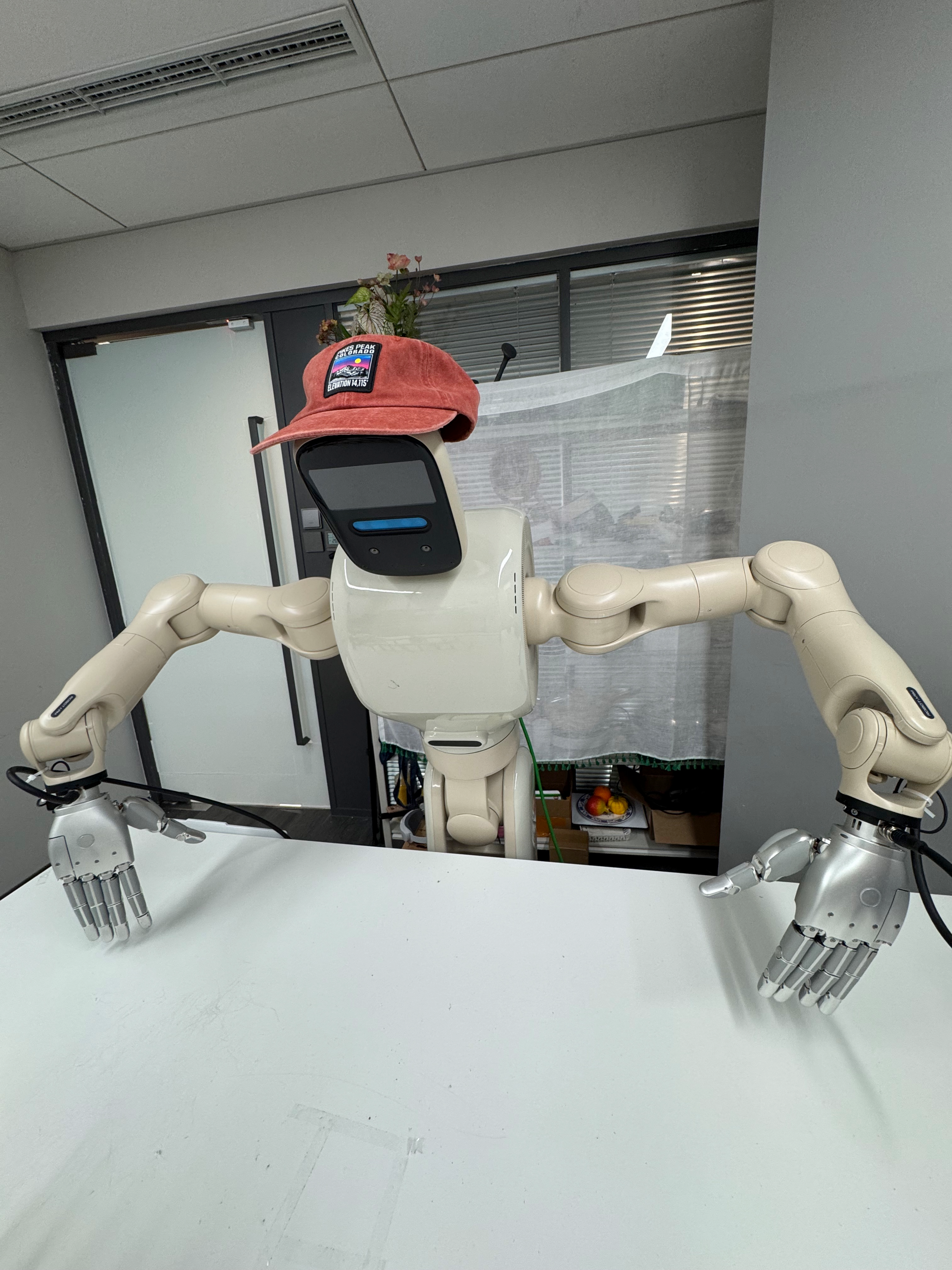}}
    \caption{\footnotesize Sharpa Wave hands with Tianji Gento Luna.}
  \end{subfigure}
  \hfill
  \begin{subfigure}[t]{0.48\linewidth}
    \centering
    \parbox[c][0.20\textheight][c]{\linewidth}{\centering
      \includegraphics[height=0.20\textheight]
      {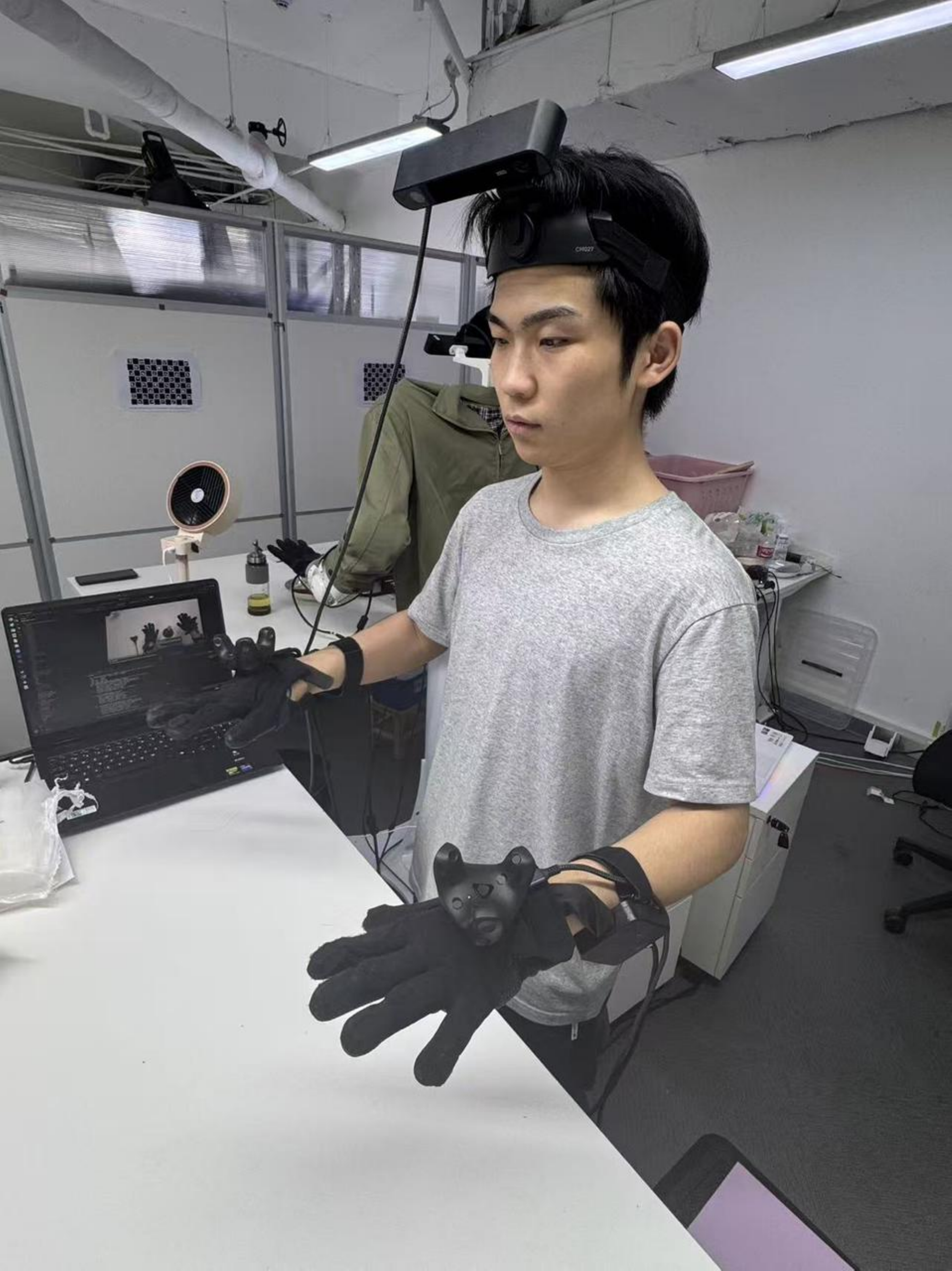}}
    \caption{\footnotesize Human--robot alignment setup.}
  \end{subfigure}
  \caption{\textbf{Robot platforms and human--robot alignment collection.}
  \textbf{(a)} Wuji hands with Tianji Marvin dual arms.
  \textbf{(b)} Wuji Hand~2 with Tianji Gento Luna.
  \textbf{(c)} Sharpa Wave hands with Tianji Gento Luna.
  \textbf{(d)} Human--robot alignment collection setup.}
  \label{fig:robot-platforms-and-human-collection}
\end{figure}

We use three bimanual robot configurations for data collection and
post-training. The first combines a Tianji Marvin dual-arm system with two
20-DoF Wuji dexterous hands. This platform, shown in
Fig.~\ref{fig:robot-platforms-and-human-collection}(a), additionally integrates ZED Mini and ZED~2
cameras and Wuji Robot Glove tactile sensing. The other two configurations both use
the Tianji Gento Luna dual-arm system: one is equipped with 20-Dof Wuji Hand 2, and the other with 22-DoF Sharpa Wave; these two
configurations are summarized in Fig.~\ref{fig:robot-platforms-and-human-collection}.

\subsection{Teleoperation}

The three systems use platform-specific control interfaces while retaining a
common separation between teleoperation, recording control, and data storage.
On the Marvin--Wuji platform, MANUS glove observations are first converted to
a 21-keypoint MediaPipe hand representation and then retargeted to the Wuji
joint space. Three VIVE Ultimate Trackers provide torso and bilateral wrist
poses. Relative wrist motion is converted into dual-arm end-effector targets,
which are solved by the Tianji arm inverse-kinematics controller.
On both Gento Luna platforms, the left and right PICO controller poses define
the dual-arm targets. Glove observations are independently retargeted to the
corresponding Wuji or Sharpa hand.

\section{Human--Robot Alignment Data Collection}
\label{app:human-robot-align}

Human--robot alignment data are collected without commanding a robot. The
operator performs manipulation tasks while wearing a pair of Wuji Human
Gloves and one HTC VIVE Tracker on each wrist. A third tracker is fixed to the
table as a stationary workspace reference. A ZED~2 stereo camera records the
first-person left and right RGB views.

The table-fixed tracker defines the reference frame used for alignment. Let
${}^{V}\mathbf{T}_{F}$ and ${}^{V}\mathbf{T}_{W_h}$ denote the fixed tracker
and wrist poses in the VIVE Lighthouse frame for hand $h\in\{L,R\}$. The
wrist pose expressed in the fixed reference frame is
\begin{equation}
    {}^{F}\mathbf{T}_{W_h}
    =
    \left({}^{V}\mathbf{T}_{F}\right)^{-1}
    {}^{V}\mathbf{T}_{W_h}.
\end{equation}
This representation removes the arbitrary Lighthouse origin and expresses
human wrist motion in a stable workspace-centered frame.

\section{Egocentric Data Curation and Offline Annotation}
\label{app:ego-data-processing}

We convert recordings with hand-pose observations into a unified
representation through pose standardization, quality-aware temporal
segmentation, and language annotation. Each processed episode contains
synchronized RGB streams, hand poses, and natural-language annotations. When
multiple views or auxiliary modalities are available, identical temporal
boundaries are applied to every signal to preserve frame-level synchronization.

\subsection{Pose Standardization}

Hand observations are represented by a 134-dimensional vector:
\begin{equation}
    \mathbf{e}_t =
    \left[
    \mathbf{p}^{L}_t,
    \mathbf{q}^{L}_t,
    \mathbf{k}^{L}_t,
    \mathbf{p}^{R}_t,
    \mathbf{q}^{R}_t,
    \mathbf{k}^{R}_t
    \right],
    \label{eq:ego-pose-vector}
\end{equation}
where $\mathbf{p}^{h}_t \in \mathbb{R}^{3}$ and
$\mathbf{q}^{h}_t \in \mathbb{R}^{4}$ denote the world-frame wrist
translation and an $xyzw$ quaternion for hand $h$, respectively.
$\mathbf{k}^{h}_t \in \mathbb{R}^{60}$ contains the wrist-local coordinates
of 20 non-wrist hand keypoints. The wrist keypoint is not repeated in
$\mathbf{k}^{h}_t$.

When wrist poses are not already expressed in a common world coordinate
system, available camera trajectories and calibration parameters transform
them into the world frame. Finger coordinates are then expressed relative to
the corresponding wrist. Original RGB streams are retained, while available
depth, inertial, calibration, camera-pose, and motion-capture signals are
preserved as synchronized auxiliary information.

\subsection{Quality-Aware Temporal Processing}

To curate long and noisy ego-centric recordings, we design a quality-aware
temporal processing pipeline, termed BoxTrim. BoxTrim removes
unreliable temporal regions and converts the remaining recordings into
compact, action-centric clips while preserving synchronization across all
available modalities.

The pipeline targets three common failure modes in ego-centric recordings:
(i) rapid body turns that induce abrupt viewpoint changes,
(ii) large translational movements or discontinuities in reconstructed hand
trajectories, and
(iii) intervals in which both hands leave the observable field. Because
reliable camera-pose signals are not consistently available for every source,
BoxTrim uses wrist-trajectory discontinuities as a unified proxy for rapid
turns and large ego-centric motion. Simultaneous invalidity of the two
hand-keypoint tracks is used as a proxy for both hands leaving the observable
field or for complete hand-tracking failure.

Let $\mathbf{p}^{L}_t$ and $\mathbf{p}^{R}_t$ denote the reconstructed
positions of the left and right wrists. We define the instantaneous motion
magnitude as
\begin{equation}
    d_t =
    \max\left(
        \left\lVert
            \mathbf{p}^{L}_{t+1}-\mathbf{p}^{L}_{t}
        \right\rVert_2,
        \left\lVert
            \mathbf{p}^{R}_{t+1}-\mathbf{p}^{R}_{t}
        \right\rVert_2
    \right).
    \label{eq:ego-wrist-motion}
\end{equation}
An adaptive threshold is estimated from the empirical wrist-motion
distribution of each recording:
\begin{equation}
    \tau_{\mathrm{motion}} =
    \max\left(
        \alpha\,
        \operatorname{median}\{d_t \mid d_t>0\},
        \tau_{\min}
    \right),
    \label{eq:ego-adaptive-motion-threshold}
\end{equation}
where $\alpha$ is a robust scale factor and $\tau_{\min}$ prevents nearly
static recordings from producing an overly sensitive threshold. Frames
satisfying $d_t>\tau_{\mathrm{motion}}$ are treated as motion
discontinuities. A short temporal margin is added on both sides of every
detected event to remove neighboring frames that may also contain corrupted
pose estimates.

To identify intervals in which the hands are not reliably observed, let
$\mathbf{k}^{h}_t$ denote the finger-keypoint vector of hand $h$. We define
\begin{equation}
    z^h_t =
    \mathbb{I}\!\left[
        \left\lVert\mathbf{k}^{h}_t\right\rVert_1 < \epsilon
        \;\lor\;
        \neg\operatorname{Finite}(\mathbf{k}^{h}_t)
    \right],
    \label{eq:ego-invalid-hand-track}
\end{equation}
and mark a frame as unreliable when both hand tracks are invalid:
\begin{equation}
    b_t = z^L_t \land z^R_t.
    \label{eq:ego-both-hands-invalid}
\end{equation}
Sustained runs of $b_t=1$ are interpreted as intervals in which both hands
have left the observable field or the hand tracker has otherwise failed.
This pose-space criterion does not require an explicit image-plane hand
detector.

The high-motion and invalid-hand intervals are merged and removed from the
original recording. The same temporal mask is applied to all synchronized
RGB streams, pose trajectories, depth observations, calibration data, and
auxiliary motion signals. When semantic temporal annotations are available,
their timestamps are first mapped to video-frame indices and used as
preferred clip boundaries. Overlapping intervals are merged, and long
intervals are further divided near low-motion frames. Very short fragments
are discarded, while long clips are subdivided into more balanced temporal
chunks. Residual non-finite values are sanitized during a final consistency
pass, and all resulting clips are deterministically renumbered.

The exact temporal thresholds are selected according to the frame rate,
annotation density, and tracking characteristics of each data source. We do not assume a universal frame rate or apply a single set of frame-count
thresholds to all corpora.

\subsection{Language Annotation}

After conversion, each clip is annotated with a vision--language model. For
recordings with multiple synchronized views, one consistently selected
egocentric RGB stream is used for caption generation while all views remain
in the processed data. The model returns structured descriptions of
(i) the environment, (ii) the overall manipulation behavior, and
(iii) left--right differentiated hand behavior.

\section{Hand Retargeting Algorithm}
\label{app:retargeting}

The egocentric human data used for pretraining represent each hand using 21
MediaPipe-format 3D keypoints. We convert these keypoints offline to the
20-DoF joint space of the first-generation Wuji hand. Let
$\mathbf{X}_{t}^{h}\in\mathbb{R}^{21\times 3}$ denote the keypoints of hand
$h\in\{L,R\}$ at time $t$, with keypoint~0 corresponding to the wrist. We
first express all keypoints relative to the wrist. A wrist-centered hand frame
is then estimated from the wrist, index-finger MCP, and middle-finger MCP
landmarks. Specifically, the palm normal is estimated by singular value
decomposition, and the remaining axes are constructed from the
wrist-to-middle-MCP direction and the palm normal. A side-specific rigid
rotation maps the left- and right-hand observations into the corresponding
Wuji model coordinates.

For each valid frame, we solve an inverse-kinematics problem using the Wuji
URDF model. Let $\mathbf{q}\in\mathbb{R}^{20}$ denote a candidate Wuji
joint-angle vector, $\mathbf{q}_{t}^{*}$ the optimized joint-angle vector at
time $t$, and $\mathbf{q}_{\min},\mathbf{q}_{\max}\in\mathbb{R}^{20}$ the
joint-limit vectors. We define the feasible joint space as
$\mathcal{Q}=\{\mathbf{q}\mid
\mathbf{q}_{\min}\leq\mathbf{q}\leq\mathbf{q}_{\max}\}$, where the
inequalities are applied elementwise. Let $\mathcal{F}$ denote the set of
five fingers. For compactness, we write
$\mathcal{L}_{\mathrm{tip},t}^{f}(\mathbf{q})
=\mathcal{L}_{\mathrm{tip}}^{f}(\mathbf{q},\mathbf{X}_{t}^{h})$
and
$\mathcal{L}_{\mathrm{full},t}^{f}(\mathbf{q})
=\mathcal{L}_{\mathrm{full}}^{f}(\mathbf{q},\mathbf{X}_{t}^{h})$.
The implemented objective is
\begin{equation}
    \mathbf{q}^{*}_{t}
    =
    \arg\min_{\mathbf{q}\in\mathcal{Q}}
    \sum_{f\in\mathcal{F}}
    \left[
        \alpha_{t,f}\mathcal{L}_{\mathrm{tip},t}^{f}(\mathbf{q})
        +
        (1-\alpha_{t,f})\mathcal{L}_{\mathrm{full},t}^{f}(\mathbf{q})
    \right]
    +
    \lambda
    \left\lVert
        \mathbf{q}-\mathbf{q}^{*}_{t-1}
    \right\rVert_{2}^{2}.
    \label{eq:wuji-hand-retargeting}
\end{equation}

The objective combines two losses with different geometric roles. The full-finger loss
$\mathcal{L}_{\mathrm{full},t}^{f}$ preserves the overall configuration of
finger $f$. It compares three vectors measured from the human wrist with
their robot counterparts measured from the robot palm: the vectors to the
proximal joint, the distal joint, and the fingertip. For the four non-thumb
fingers, these points correspond to the PIP, DIP, and fingertip landmarks;
for the thumb, the MCP, IP, and fingertip landmarks are used. The human
vectors are rescaled to account for differences between human and robot
finger lengths, while the robot vectors are computed through forward
kinematics. The loss is the average Huber penalty over the three vector
discrepancies.

The fingertip loss $\mathcal{L}_{\mathrm{tip},t}^{f}$ places greater
emphasis on the fingertip geometry. It compares both the fingertip position
relative to the wrist or robot palm and the direction of the final finger
segment. The human segment direction is measured from the DIP landmark to
the fingertip, or from the IP landmark to the fingertip for the thumb. The
corresponding robot direction is obtained from the final finger link through
forward kinematics. Huber penalties on the fingertip-position and direction
discrepancies are combined to form
$\mathcal{L}_{\mathrm{tip},t}^{f}$.

The interpolation weight $\alpha_{t,f}$ determines the relative importance
of the two losses. For each non-thumb finger, it is computed from the distance
between that fingertip and the thumb tip. As the two fingertips approach,
$\alpha_{t,f}$ increases and places more emphasis on the fingertip loss.
Otherwise, the full-finger loss receives more weight and preserves the
overall finger configuration. The thumb weight is set to the maximum weight
among the other four fingers. The weights are clipped to a bounded range so
that both losses remain active. The coefficient $\lambda$ controls temporal
regularization toward the previous solution $\mathbf{q}_{t-1}^{*}$.

Non-finite or geometrically degenerate keypoint frames are removed before
optimization. The optimization is constrained by the Wuji joint limits, and
a first-order low-pass filter is applied to the optimized trajectory. The
result contains 20 finger joint angles per hand and forms the hand-joint
component of the pretraining action representation.

This retargeting procedure is applied only to human keypoint data used for
pretraining. For post-training data collected with the 20-DoF Wuji Hand~2
and the 22-DoF Sharpa Wave hands, we directly use the joint angles recorded
by the corresponding robot data-collection stacks in their native joint
spaces. These joint angles are not produced by the MediaPipe-to-Wuji
retargeting procedure described above.

\section{Model and Training Details}
\label{app:model-training}

\paragraph{Notation.}
A \emph{latent frame} is the spatial video representation produced by the
video encoder at one latent timestep. We use $W$ for the total number of latent
frames in one sampled training window and $\kappa$ for the number of consecutive
latent frames grouped into one prediction chunk; the aligned action segment is
grouped into the same chunk. A history width measured in chunks therefore
corresponds to $\kappa$ times as many latent frames.

\subsection{Pre-training}

\paragraph{Stage 1: video-only pre-training.}
We initialize the visual backbone from Wan~2.2-TI2V-5B and train only the video
pathway with conditional flow matching. Stage~1 first uses low-resolution
monocular egocentric clips for 500K steps and then high-resolution monocular
clips for 340K steps. Each sample provides one or two clean latent frames as
context. Action and value tokens are absent, and the later action-first mask is
not used.

\paragraph{Stage 2: joint video--action pre-training.}
Stage~2 starts from the video-only checkpoint and trains for 450K steps on
stereo egocentric video paired with human actions. Stereo views share temporal
and vertical RoPE coordinates and use offset horizontal ranges. Video and action
tokens are mutually visible within a chunk and are jointly supervised by flow
matching; the value loss remains disabled. The maximum training window contains
$W=12$ latent frames, each prediction chunk contains $\kappa=2$ frames, and the
visible clean-history width is sampled uniformly from one to five chunks. At
streaming inference the history width is fixed to two chunks. Stage~2 uses
AdamW with learning rate $2\times10^{-5}$, weight decay $0.01$, a constant
schedule with 1K warm-up steps, gradient clipping at $0.5$, and bfloat16
training.

\subsection{Mid-training}

Mid-training adapts the Stage~2 checkpoint to the robot domain using more than
100 hours of data: robot trajectories supplemented with human--robot alignment
data. We switch from the
joint within-chunk mask to the action-first mask and mix the policy,
simulation, and evaluation modes in \S\ref{sec:unified-world-model}. The
robot-domain recipe uses $640\times384$ stereo inputs, $W=8$ latent frames,
$\kappa=2$ frames per chunk, a training history of one to three chunks, and a
two-chunk streaming cache. Simulation- and evaluation-mode samples account for
$0.10$ and $0.05$ of training samples, respectively; the remaining samples use
policy mode. The value target is discretized into 201 bins with loss weight
$0.05$. We use AdamW with learning rate $2\times10^{-5}$, weight decay $0.01$,
200 warm-up steps, a linear schedule, gradient clipping at $0.5$, and bfloat16
training.

\subsection{Post-training}

Post-training starts from the mid-trained checkpoint and specializes the model
to target-robot control. Supervised adaptation retains the action-first mask;
the branches below change the temporal context or add the self-evolution and
tactile objectives.

\paragraph{Sliding-window SFT.}
The default target-robot policy uses $W=8$, $\kappa=2$, a training history
sampled from one to three chunks, and a two-chunk streaming cache. Temporal RoPE
is window-relative, so surviving video keys are rebased after cache eviction.
Action chunks are sampled with five flow-matching denoising steps during default
deployment.

\paragraph{Global autoregression.}
The full-history branch trains on complete episodes with global temporal RoPE
and a video-shared latent clock for action tokens. It disables compressed memory
by setting the numbers of anchor and memory tokens to zero and makes the recent
window non-binding, so every prediction chunk can attend to all preceding clean
observations. This branch therefore retains $NL$ visual KV tokens after $N$
latent frames.

\paragraph{Hybrid working memory.}
Our MemoryWAM adaptation uses two full-resolution anchor frames, four
full-resolution recent frames, and eight memory tokens per latent frame. It is
trained on complete episodes with global temporal RoPE and video-shared action
positions. In the reported long-context comparison, the global-autoregressive
and hybrid branches use 10K optimization steps, learning rate
$2\times10^{-5}$, AdamW betas $(0.9,0.95)$, weight decay $0.01$, 200 warm-up
steps, gradient clipping at $0.5$, and 50 action-denoising steps for evaluation.
Their variable-length loader uses a latent-frame budget of 100 and packs at most
eight complete episodes per micro-batch; an episode exceeding the budget is
kept intact as a singleton rather than truncated.

\paragraph{MBRL and planning.}
For each sampled prefix, the rollout pipeline generates eight policy candidates
and adds one ground-truth candidate. Prefixes contain one to four chunks and the
imagined rollout horizon is one chunk. In the asynchronous joint-rollout
pipeline, action and future-video sampling use four and eight denoising steps,
respectively. DiffusionNFT uses $\beta=0.1$ and one sampled noise level per
candidate. The EMA reference has decay $0.99$ and is published every ten trainer
updates. Only action-related parameters are optimized; the video backbone and
value components remain frozen. Rollouts from stale reference versions or more
than 600 seconds old are discarded.

\paragraph{Tactile expert.}
We use a common tactile-expert architecture and temporal schedule for the
Sharpa Wave and Wuji Hand~2 dexterous hands. The expert is a 30-layer
transformer with hidden width 128, four attention heads, and feed-forward width
512. Its layers are paired one-to-one with the shared backbone: each layer
self-attends over tactile and intermediate-action tokens and reads the cached
K/V states from the corresponding backbone layer. The backbone denoises each
48-action chunk to $\sigma_c=0.2$ once, then refreshes the action K/V at this
intermediate state and concatenates them with the visible video-context K/V.
The resulting layer-wise cache is detached and reused by the tactile expert.
At the action rate of 30~Hz, the chunk is divided into eight six-action
sub-chunks, each spanning 0.2 seconds. The expert refines these sub-chunks
sequentially using the immediately preceding tactile window. Force signals are
sampled at 90~Hz, while the Sharpa Wave additionally provides six deformation
frames per window. The same intermediate action chunk and KV cache are reused
across all eight calls. The intermediate actions and cached states are detached so
that the tactile losses update only the expert, the force encoder and decoder,
and, for the Sharpa Wave, the deformation encoder.

\emph{Sharpa Wave.}
The expert receives a 60-dimensional bimanual force--torque vector together
with deformation images from ten finger pads. A temporal convolutional encoder
maps six consecutive deformation frames to 20 conditioning tokens. The
deformation images are not prediction targets.

\emph{Wuji Hand~2.}
The expert receives a 1086-dimensional bimanual vector containing 1056
pointwise force values from 176 three-axis sensing points per hand and 30
per-finger aggregate three-axis force values.

For both dexterous hands, force inputs are normalized using the configured
training-data statistics and invalid channels are set to zero. For force
prediction, the immediately preceding force window is encoded without noise
and used as a conditioning state; the Sharpa Wave additionally conditions on
the preceding deformation images. The real following force window is corrupted
in normalized signal space and passed through the same force encoder. The
expert performs conditional flow matching over the resulting tokens, and a
force decoder maps the future-force token states back to a velocity in force
space. The prediction loss is evaluated only against the flow target
constructed from the real following force signal. We jointly optimize the
expert, force encoder and decoder, and the Sharpa deformation encoder using
Eqs.~(\ref{eq:tactile-refinement-loss}) and
(\ref{eq:tactile-prediction-loss}), with $\lambda_{\mathrm{pred}}=0.1$.
Optimization uses AdamW with a learning rate of $5\times10^{-5}$ and weight
decay $0.01$.

\section{Qualitative Value-Model Trajectories}
\label{app:value-model-visualizations}

Figures~\ref{fig:value-model-trajectories}
and~\ref{fig:value-model-trajectories-b} show predicted task progress along
recorded trajectories from four tasks that differ in horizon, in the manipulated
object, and in their failure modes. Each task contributes one successful and one
failed execution, and all eight panels are scored by the same value model.

\paragraph{Protocol.}
Values are read out in the evaluation mode of Eq.~(\ref{eq:uwm-modes}), so the
prediction for chunk~$t$ is conditioned on the context, the executed actions and
the observed video of that chunk, and never on the outcome: whether the episode
eventually succeeded is not provided to the model. Each recorded
episode is replayed through the streaming window used during training, and one
value is read per latent frame rather than per video frame, which sets the
spacing of the plotted points. Values follow the
relative-progress reward of Eq.~(\ref{eq:relative-progress-reward}) and
Eq.~(\ref{eq:random-negative-relative-reward}) and therefore lie in $[-1,1]$;
they are trained to rank rollouts relative to one another and are not calibrated
success probabilities. Keyframes above each panel are timestamped
and show the left-camera view only, and all eight panels share the same vertical
scale.

\paragraph{Successful executions.}
Panels (a), (c), (e) and (g) terminate between $0.63$ and $0.65$ but follow
distinct profiles. On Cut Reeds the prediction remains near zero for the first
15~s, during which the hand approaches and aligns but no reed has been cut, and
then increases in two steps as the cuts are completed. On Make Tea, Multi-Finger
and Put Phone the prediction increases smoothly from the outset,
consistent with tasks in which most of the motion contributes incrementally to
completion. In all four panels the predicted level alone indicates how far the
task has progressed.

\paragraph{Failed executions.}
The failed executions are not uniformly low, and their shape is the more
informative signal. On Cut Reeds (b) the prediction follows its successful
counterpart into the cutting phase and reaches $0.27$. The first cut then fails
to sever the reed and lifts it instead, and the prediction falls to $-0.39$
within two seconds; it declines further to $-0.51$ as the rack holding the reeds
is knocked out of alignment. Put Phone (h) exhibits the largest
excursion: it peaks at $0.60$ once the phone has been placed inside the box, and
decreases to $-0.26$ only after the robot has repeatedly failed to close the
lid, which the task requires. On Make Tea (d) the prediction rises to $0.14$ at
approximately 35~s, by which point the water has been poured into the cup
holding the tea leaves, and then decreases over the remaining 80~s to $-0.47$ as
the robot fails to lift the lid off the table. Multi-Finger (f) rises to
$0.18$ near 50~s and falls to $-0.53$ at the end of the episode: the task
requires the objects held in the multi-finger grasp to be placed in separate
containers, and here they are all released into the same one. This common
structure---an increase while the early portion of the task is executed
correctly, followed by a decrease once progress stops---is what allows the value
model to distinguish a recoverable state from an unrecoverable one when MBRL
scores imagined rollouts.

% One logical figure over two pages.  [p] gives each half its own float page,
% so the section text is not left stranded above a figure too tall to join it.
% \ContinuedFloat holds the counter; the local \thefigure adds the a/b suffix,
% which \ContinuedFloat* does not produce under this class.
\begin{figure}[p]
  \renewcommand{\thefigure}{\arabic{figure}a}
  \centering
  \includegraphics[width=\linewidth]{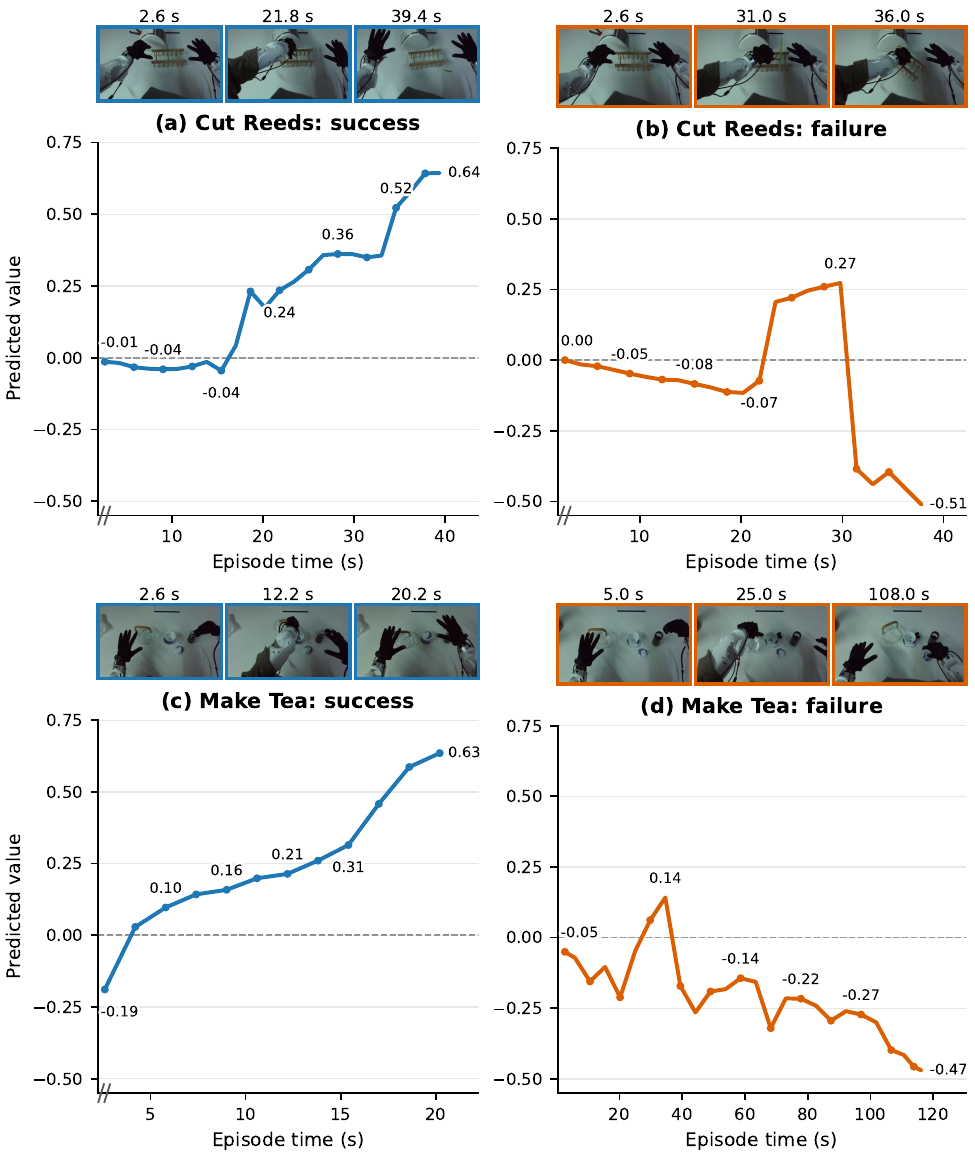}
  \caption{\textbf{Predicted task progress: Cut Reeds and Make Tea.}}
  \label{fig:value-model-trajectories}
\end{figure}

\begin{figure}[p]
  \ContinuedFloat
  \renewcommand{\thefigure}{\arabic{figure}b}
  \centering
  \includegraphics[width=\linewidth]{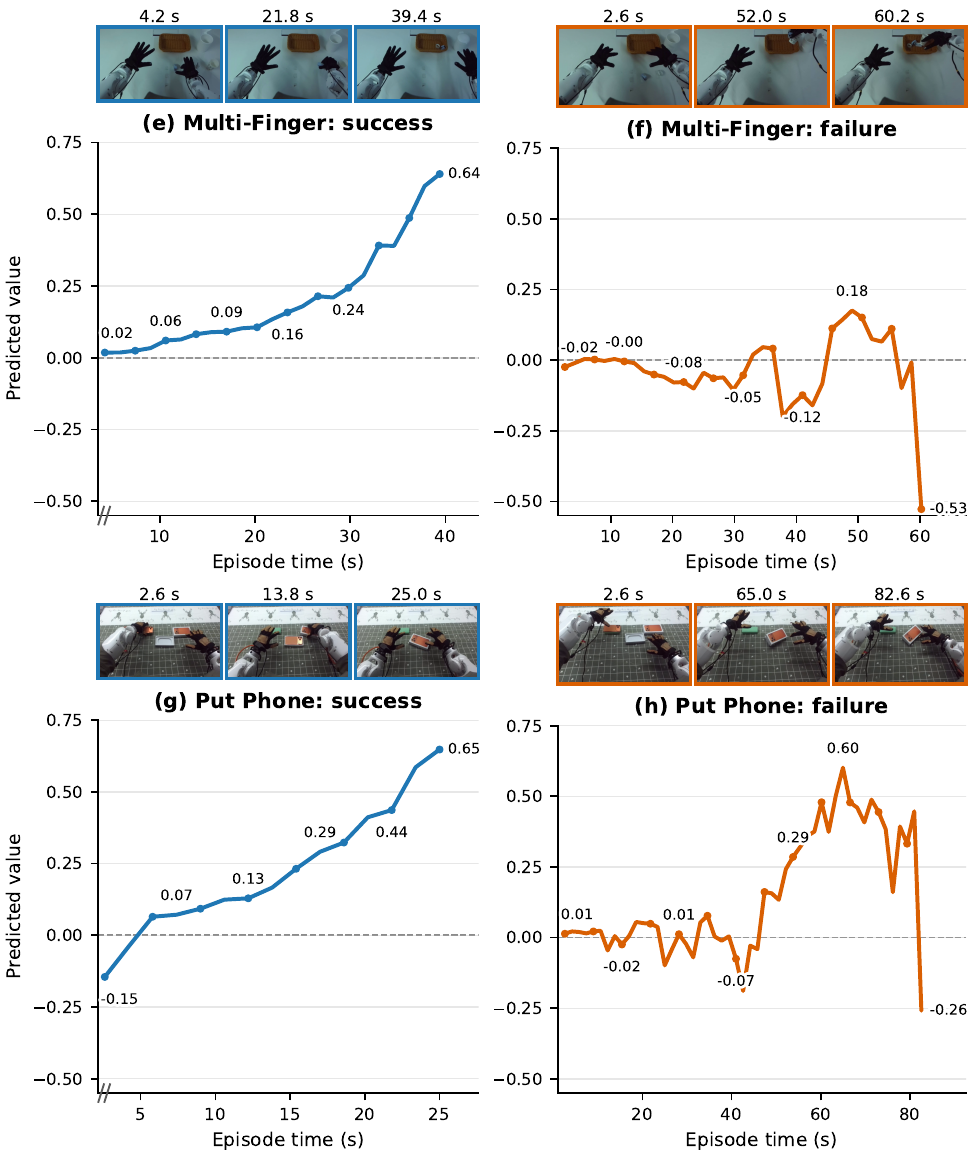}
  \caption{\textbf{Predicted task progress: Multi-Finger and Put Phone.}}
  \label{fig:value-model-trajectories-b}
\end{figure}

\clearpage
\section{Evaluation Tasks}
\label{app:evaluation-tasks}

\begin{figure}[H]
  \centering
  \setlength{\abovecaptionskip}{4pt}

  \includegraphics[width=\linewidth]{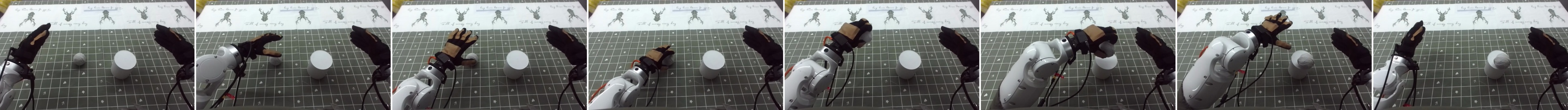}
  \parbox{\linewidth}{\footnotesize\textbf{(a) Place Ball.} The robot uses its
  left hand to pick up a ball and place it on top of an inverted cup.}
  \par\vspace{2pt}

  \includegraphics[width=\linewidth]{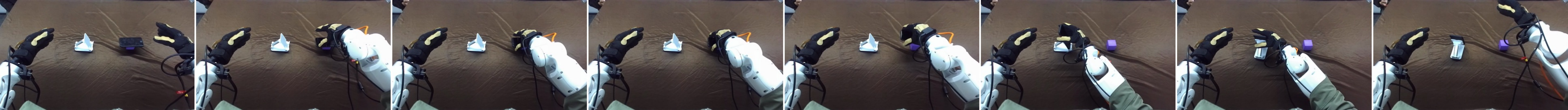}
  \parbox{\linewidth}{\footnotesize\textbf{(b) Put Phone.} The robot uses its
  right hand to pick up a smartphone from a purple block and place it into a
  white phone stand.}
  \par\vspace{2pt}

  \includegraphics[width=\linewidth]{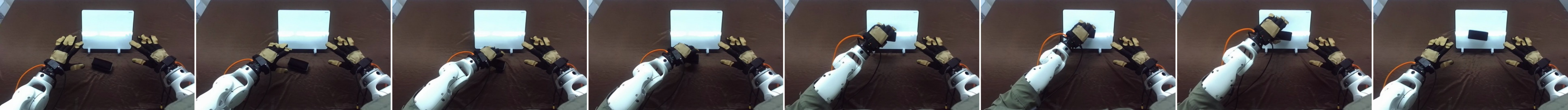}
  \parbox{\linewidth}{\footnotesize\textbf{(c) Attach Eraser.} The robot uses
  its left hand to pick up an eraser and attach it to the whiteboard surface.}
  \par\vspace{2pt}

  \includegraphics[width=\linewidth]{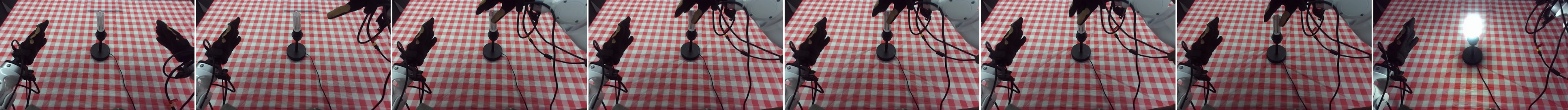}
  \parbox{\linewidth}{\footnotesize\textbf{(d) Screw Bulb.} The robot uses its
  right hand to screw a bulb into a socket until it lights, then releases it.}
  \par\vspace{2pt}

  \includegraphics[width=\linewidth]{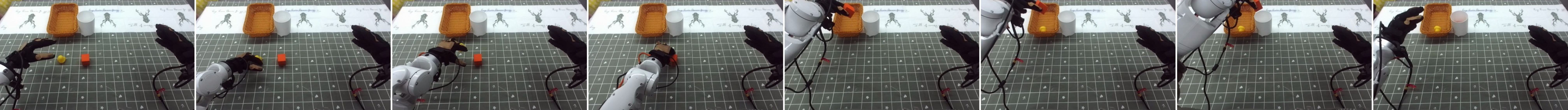}
  \parbox{\linewidth}{\footnotesize\textbf{(e) Multi-Finger.} The robot grasps
  an object from a set of differently shaped items and places it into an orange
  basket.}
  \par\vspace{2pt}

  \includegraphics[width=\linewidth]{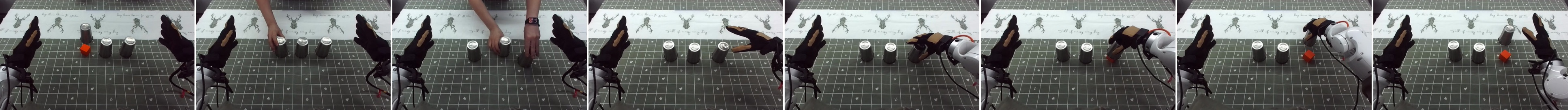}
  \par\vspace{1.5pt}
  \parbox{\linewidth}{\footnotesize\textbf{(f) Find Square.} After a person
  hides an orange square under one of three cups and shuffles them, the robot
  must use the earlier observation to lift the correct cup.}
  \par\vspace{2pt}

  \includegraphics[width=\linewidth]{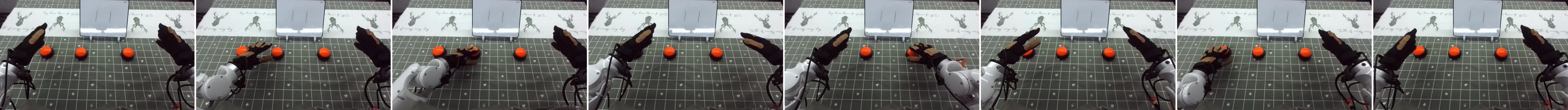}
  \parbox{\linewidth}{\footnotesize\textbf{(g) Press Button.} The robot observes
  a visual cue, retains it while acting, and presses the indicated physical
  button sequence.}
  \par\vspace{2pt}

  \includegraphics[width=\linewidth]{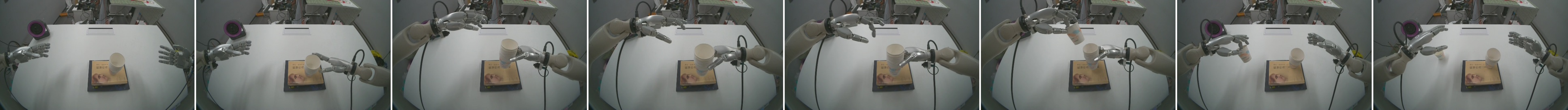}
  \parbox{\linewidth}{\footnotesize\textbf{(h) Pull Out Paper Cup.} The robot
  stabilizes a nested stack and extracts a single paper cup without dropping or
  crushing it.}
  \par\vspace{2pt}

  \includegraphics[width=\linewidth]{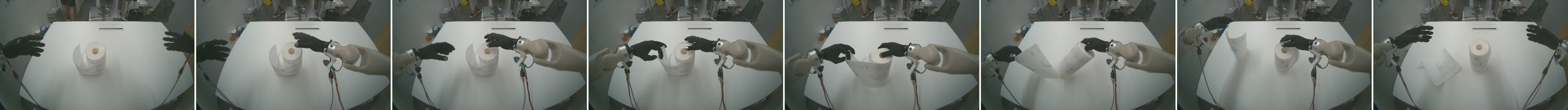}
  \parbox{\linewidth}{\footnotesize\textbf{(i) Tear Paper.} The robot
  coordinates both hands to grasp and tear a sheet of paper from the roll.}

  \caption{\textbf{Evaluation tasks.} Each row shows eight frames sampled from
  a representative episode. \textbf{(a)--(e)} are the main-suite tasks,
  \textbf{(f)--(g)} are the working-memory tasks, and \textbf{(h)--(i)} are the
  tactile tasks. The MBRL study reuses Put Phone and Multi-Finger from the main
  suite under different policy-optimization and inference settings.}
  \label{fig:evaluation-tasks}
\end{figure}

Figure~\ref{fig:evaluation-tasks} summarizes the nine unique physical tasks.
The MBRL study reuses two tasks from the main suite rather than introducing
additional task definitions.

\end{document}